\documentclass[smallextended]{svjour3}       

\usepackage{authblk}
\usepackage[utf8]{inputenc}  
\usepackage[T1]{fontenc}   
\usepackage{graphicx}
\usepackage{xcolor}
\usepackage{amsmath,amssymb,amsfonts}
\usepackage{multirow}
\usepackage{cite}

\usepackage[boxed]{algorithm2e}

\journalname{}
\def\makeheadbox{\relax}

\begin{document}

\title{Fractional Naive Bayes (FNB):\\ non-convex optimization for a parsimonious\\ weighted selective naive Bayes classifier}
\titlerunning{A parsimonious weighted selective naive Bayes classifier}

\author{Carine Hue \and Marc Boullé} 

\institute{C. Hue and M. Boullé \at
              Orange Labs - 22300 Lannion - France \\
              \email{firstname.name@orange.com}
}

\date{September, 2024}

\maketitle

\begin{abstract}
We study supervised classification for datasets with a very large number of input variables. The naïve Bayes classifier is attractive for its simplicity, scalability and effectiveness in many real data applications.
When the strong naïve Bayes assumption of conditional independence of the input variables given the target variable is not valid, variable selection and model averaging are two common ways to improve the performance.
In the case of the naïve Bayes classifier, the resulting weighting scheme on the models reduces to a weighting scheme on the variables. Here we focus on direct estimation of variable weights in such a weighted naïve Bayes classifier. We propose a sparse regularization of the model log-likelihood, which takes into account prior penalization costs related to each input variable. Compared to averaging based classifiers used up until now, our main goal is to obtain parsimonious robust models with less variables and equivalent performance.
The direct estimation of the variable weights amounts to a non-convex optimization problem for which we propose and compare several two-stage algorithms. First, the criterion obtained by convex relaxation is minimized using several variants of standard gradient methods. Then, the initial non-convex optimization problem is solved using local optimization methods initialized with the result of the first stage. The various proposed algorithms result in optimization-based  weighted naïve Bayes classifiers, that are evaluated on benchmark datasets and  positioned w.r.t. to a reference averaging-based classifier.

\end{abstract}

\section{Introduction}

Due to the continuous increase of storage capacities, data acquisition and processing have deeply evolved during the last decades. Henceforth, it is common to process datasets including a very large number of variables. With an automatic variable construction approach like in \cite{BoulleEtAlML18}, the space of variables that can be constructed is even virtually infinite and variables can be highly correlated.\\
In this context, we consider the supervised classification problem where $Y$ is a categorical target variable 
with $J$ values $C_1,\ldots,C_J$ and $X = (X_1,\ldots,X_K)$ is the set of $K$ input variables, numerical or categorical. 

In this article we focus on the naïve Bayes classifier \cite{Langley:1992} which is attractive for its simplicity, scalability and effectiveness in many real data applications \cite{Hand:2001}.
The naive Bayes classifier is a simple probabilistic classifier based on applying Bayes' theorem with naive conditional independence assumption. The input variables $(X_k)_{k=1,\ldots,K}$ are assumed to be independent given the target variable $Y$. Despite this strong assumption, this classifier has proved to be effective in many real applications, e.g. in text classification \cite{mccallum1998naive}, for medical diagnosis \cite{Wolfson:2015}, on data stream \cite{Gama:2010}. Some conditions of its optimality have also been studied in  \cite{Zhang2004} \cite{Domingos:1997}.

This "naïve" assumption allows us to compute the model directly from the univariate conditional estimates $P(X_{k}|C)$. For an instance denoted $n$, the probability of the target value $C$ conditionally to the value of the input variables is computed according to the formula \footnote{We consider in this paper that estimates of prior probabilities  $P(Y=C_j)$ and of conditional probabilities  $p(x_k|C_j)$ are available. In our experiments, these probabilities are estimated using univariate discretization or grouping preprocessing according to the MODL method  (see resp. \cite{Boulle:2006} \cite{Boulle:2005})}:

\begin{equation}
P(Y=C|X=x^{n})=\frac{P(Y=C)\prod_{k=1}^K {p(x_k^{n}|C)}}{\sum_{j=1}^J P(Y=C_j)\prod_{k=1}^K {p(x_k^{n}|C_j)}}
\label{LL}
\end{equation}

The naive independence assumption can harm the performance when violated. In order to better
deal with highly correlated variables, the selective naive Bayes approach \cite{Langley:1994}
exploits a wrapper approach \cite{Kohavi:1997} to select the subset of variables which
optimizes the classification accuracy.
Another way to improve the learned classifiers is to train multiple models, in our case several selective naive Bayes classifiers with different subsets of variables, and combine them using model averaging (see for example \cite{Hoeting:1999}).
Moreover Boull\'e in \cite{Boulle:2006:b} shows the close relation between weighting variables and averaging selective naive Bayes classifiers in the sense that, in the end, the two processes produce a similar single model where a weight is given to each input variable. Equation \ref{LL} is just turned into the following equation with a weight $w_k \in [0,1]$ per variable $X_k$:

\begin{equation}
P_w(Y=C|X=x^{n})=\frac{P(Y=C)\prod_{k=1}^K {p(x_k^{n}|C)^{\textcolor{red}{w_k}}}}{\sum_{j=1}^J P(Y=C_j)\prod_{k=1}^K {p(x_k^{n}|C_j)^{\textcolor{red}{w_k}}}}
\label{LL2}
\end{equation}

In this paper, we  particularly focus on weighing variables for datasets with a very large number of variables, in the order of $10000$.
One of the advantages of the classifier described by Equation (\ref{LL2}) is its low complexity for model deployment, which is linear with the number of input variables: a weighted naive Bayes classifier is completely described by its weight vector $w =(w_1,w_2,\ldots,w_K)$. The interpretation of the learned models is also simpler than in the case of ensembles of models.\\
Within the \emph{weighted naive Bayes classifier} family, we can distinguish:
\begin{itemize}
\item[-]{classifiers with all weights equal to $1$. It corresponds to the standard naïve Bayes classifier that uses all the input variables.}
\item[-]{classifiers with Boolean weights in $\{0,1\}^K$. It corresponds to the selective naïve Bayes classifiers that selects a subset of input variables.
}
\item[-]{classifiers with continuous weights in $[0,1]^K$. Such classifiers can be obtained by averaging classifiers with Boolean weights with a weighting proportional to the posterior probability of the model \cite{Hoeting:1999} or proportional to their compression rate \cite{Boulle:2007b}.
}
\end{itemize}

For datasets with a very large number of variables, models derived from such an averaging approach retain most of the variables, as they are used by one of the averaged selective naive Bayes classifiers, resulting in non-zero weights. This makes the obtained classifiers both costly to deploy and difficult to interpret.
Note that this Selective Naïve Bayes (SNB) predictor based on variable selection and model averaging was implemented in early versions of the tool named Khiops \cite{Khiops9}, which performs automatic feature construction for multi-table datasets, data preprocessing based on supervised discretization or value grouping, and supervised classification \cite{Boulle:2007b} or regression \cite{HueEtAlJMLR07}.
In this paper, we investigate on a new predictor named Weight Naive Bayes (WNB), where the variable weights result from a direct optimization, and we expect some of the weights to receive a zero value.
As the number of explicative variables is growing more and more in the analyzed datasets, the goal of our work is to study this WNB predictor \footnote{Implemented in the last version of the Khiops tool \cite{Khiops10}.} and compare it to the SNB predictor w.r.t. a variety of criteria, such as parsimony, performance and scalability. Our main expectation is to obtain parsimonious robust models with fewer variables while keeping at least equivalent performance as those previously obtained using the SNB predictor. 

The rest of the paper is organized as follows. A short description of the averaging approach and the current SNB are presented in Section~\ref{Averaging}. In Section~\ref{Direct}, we propose a direct estimation of the weight vector using non convex continuous optimization algorithms to minimize the regularized log-likelihood in the $[0,1]^K$ parameter space. An experimental evaluation of the obtained models in terms of parsimony, predictive performance, robustness and running times is presented in Section~\ref{Exp}, before the conclusion and future work.

\section{The Selective Naïve Bayes predictor based on an averaging approach}
\label{Averaging}

Let us present the averaging approach adopted to obtain a Selective Naïve Bayes predictor with continuous weights in $[0, 1]$.
More details can be found in \cite{Boulle:2007b}.
In this approach, we consider the space of Naïve Bayes classifiers with Boolean weights. This space is explored using a Fast Forward Backward algorithm \cite{Boulle:2007b} that adds or drops one variable at each step.
During the learning phase, each candidate Boolean model is evaluated through a regularized Bayesian criterion. 
 This approach is then based on the choice of two elements:
\begin{itemize}
\item{the choice of an evaluation criterion on the space of classifiers with Boolean weights,}
\item{the choice of an averaging algorithm to compute the best continuous weights from Boolean weights evaluations.}
\end{itemize}

\subsection{The choice of a regularized criterion}
\label{AveragingCriterion}

Let us denote by  $D_N=(x_n,y_n)_{n=1}^N$ the considered dataset. We deal with both tabular data  and data issued from relational databases. For tabular data, the explicative variables are the native variables present in the tabular representation. As real data usually comes from relational databases, we also address variables issued from an automatic construction process that efficiently explores the space of all variables that can be constructed. This process is described in \cite{BoulleEtAlML18}. The number of explicative variables that can be native or constructed is denoted by $K$.\\
The classification models we are interested in belong to the \emph{Boolean naive Bayes classifier} family. A given classification model denoted $M_w$ is described by its weight vector $w =(w_1,w_2,\ldots,w_K) \in  \{0,1\}^K$.
The fit between the data and the model can be evaluated through the negative log-likelihood of the data. Using the instance likelihood defined in (\ref{LL2}), the negative log-likelihood of the data is given by : 
\begin{equation}
F_N(w) \overset{\textrm{def}}{=} - \sum_{n=1}^N LL_n(w), w \in \{0,1\}^K,
\label{Fn}
\end{equation}
with
{\small
\begin{eqnarray}
LL_n(w) &\overset{\textrm{def}}{=}&\log{P(Y=y^{n})} + \sum_{k=1}^K {\log{p(x_k^{n}|y^{n})^{\textcolor{red}{w_k}}}} \\
&&- \log{\left( \sum_{j=1}^J P(Y=C_j)\prod_{k=1}^K {p(x_k^{n}|C_j)^{\textcolor{red}{w_k}}} \right)}
\label{LLn}
\end{eqnarray}
}
Let notice that, as $w_k$ takes Boolean values, the term ${\log{p(x_k^{n}|y^{n})^{\textcolor{red}{w_k}}}}$ reduces to $0$ if $w_k=0$  and the factor $p(x_k^{n}|C_j)^{\textcolor{red}{w_k}}$ reduces to $1$ if $w_k=0$.
The minimization of the negative log likelihood conducts to an over-fitting problem : very "complex" classifiers, i.e. with a lot of non-zero weights can exhibit a good fit of the trained data but will not be robust on new data, as well as being costly to deploy. 
The optimization of the log-likelihood is regularized using an additional regularization term, also called penalization or prior term, which expresses the prior cost of the model.
The full regularized criterion can be written: 
\begin{equation}
CR(w)= - \sum_{n=1}^N LL_n(w) + \lambda f_B(w)
\end{equation}
where  $LL_n$ refers to the log-likelihood,  $f_B$ is the prior function for Boolean weights, and $\lambda$ is a positive regularization weight.

With $\lambda=0$, we have no regularization and a risk of over-fitting.
With $\lambda=1$, we have full regularization, which is relevant in the case where all selected input variables are conditionally independent given the target variable.
However, in the case of real datasets, the naive Bayes assumption is often violated. The use of $\lambda<1$ is equivalent to relaxing the strong naive Bayes assumption, by allowing additional variables to be incorporated that are not totally conditionally independent of the others, but which provide additional information that has a positive impact on predictive accuracy.

To derive the penalization term $f_B(w)$, we apply a Bayesian model selection approach using a prior distribution on the model parameters. The same approach was exploited in the case of the SNB approach \cite{Boulle:2007b}, by considering the number of selected variable and the subset of selected variables given this number.
In this paper, we extend this approach by using more parsimonious priors and accounting for the complexity of variable construction in the case of multi-table feature engineering \cite{BoulleEtAlML18} and the complexity of data preparation, since conditional probabilities are obtained using supervised discretization \cite{Boulle:2006} or value grouping \cite{Boulle:2005} models for each input variable.

A hierarchic prior distribution is assumed for the parameters of the model to define $f_B$ : the number of selected variables $K_s$ is chosen first, before choosing the subset of selected variables.
We do not want to make any strong assumption to encode the number of selected variable $K_s$ 
and use the universal prior for integers \cite{Rissanen:1983}, which enables an extension of the hierarchical prior presented in \cite{Boulle:2007b} to the case of a potentially infinite number of variables. The cost of the choice of a number of selected variables is noted $L^{*}(K_s)$ and follows the Rissanen distribution that is as flat as possible, with larger probabilities for small integer numbers that favors parsimonious models. 

For the choice of the subset of variables, we exploit a prior probability $p(X_i)$ of being selected per input variable $X_i$. For example, in the case of a finite set of $K$ variables, we can use $p(X_i) = 1/K$ is we assume that all variables are equiprobable.
In this paper, we exploit the Bayesian approach described in \cite{BoulleEtAlML18} to automatically obtain variable prior probabilities, using selection or construction model $M_C(X)$ in the case of multi-table feature engineering and preprocessing model $M_P(X)$ to estimate the conditional probabilities.
These prior probabilities $p(M_C(X))p(M_P(X))$ per input variable can be seen as user preferences such that, for equivalent likelihoods, the "simple" variables are preferred to "complex" ones.
A variable subset of size $K_s$ corresponds to a multinomial sample of size $K_s$ with $N_i$ draws per variable, such that $N_i=1$ if variable $X_i$ is selected and $N_i=0$ otherwise.
Overall we have $\sum_{i=1}^K {N_i} = K_s$.
We then obtain the multinomial probability 
\begin{equation}
\label{MultinomialPrior}
\frac {K_s!}{N_1!\ldots N_K!} p(X_1)^{N_1} \ldots p(X_K)^{N_K}.
\end{equation}
As $N_i!=1$ for $N_i=0$ or $N_i=1$, $p(X_i)^1 = p(X_i)$, $p(X_i)^0 = 1$, the only remaining terms in Formula~\ref{MultinomialPrior} correspond to the selected variables. 
This reduces to 
\begin{equation}
\label{MultinomialPriorSimplified}
K_s! p(X_1) \ldots p(X_{K_s}),
\end{equation}
where $p(X_k)$ is the probability of the $k^{th}$ selected variable.
Taking the negative logarithm of these multinomial probabilities, the overall prior term for $w$ is finally given by:
\begin{equation}
\label{SNBPrior}
f_B(w) = L^*(K_s) - \log {(K_s !)} + \sum_{k=1}^{K_s} B(X_k)
\end{equation}
where $K_s$ is obtained as $\sum_{k=1}^K w_k$ and $B(X_k) = -\log p(M_C(X)) -\log p(M_P(X))$ is the sum of the selection, construction and preparation cost of variable $X_k$ . As the variable weights $w_k$ are Boolean, the sum $K_s=\sum_{k=1}^K w_k$ is an integer.
Overall, $f_B(w)$ can be seen as the negative log prior probability of the model or as its coding length in the minimum description length approach \cite{Rissanen:1978}. 
It extends the approach introduced with the SNB classifier \cite{Boulle:2007b} by exploiting a more parsimonious prior, by allowing the selection among an infinite number of variables and by taking into account the complexity of the input variables with their selection, construction and preprocessing model.

\subsection{The optimization algorithm}
The regularized criterion is optimized using the Fast Forward Backward algorithm described in \cite{Boulle:2007b} and summarized in Algorithm~\ref{AlgoSNB}.
The algorithm essentially consists of several forward/backward add/drop variable passes.
At each add or drop step, the considered variable is selected/eliminated if it improves/degrades the regularized criterion value.
After each addition or deletion of a variable, the selected subset of variables defines an elementary Selective Naïve Bayes classifier with Boolean weights.
After several starts of the Fast Forward Backward algorithm, the continuous weights of the final predictor are obtained by averaging the Boolean weights obtained for all the subsets of variables encountered along the optimization trajectory.

One of the strengths of this algorithm is its low time complexity equal to $O(KN \log(KN))$ where $K$ is the number of variables and $N$ is the number of instances. Each add or drop of a variable modifies only one weight. That is why, using the additivity of the criterion, the log-likelihood can be directly updated taking into account the new weight and the previous probabilities kept into memory for the other variables.
This regularization and averaging of Selective Naïve Bayes models obtains highly competitive performance on intensive benchmarks and on prediction challenges.
Nevertheless, the averaging also results in keeping almost all the initial variables as almost each one is present at least once in the evaluated subsets of variables.

\smallskip
To counter this drawback and obtain sparser classifiers, we study in the following an alternative approach based on a direct optimization of the variable weights.

\begin{algorithm}[h]
\caption{Algorithm MS(FFWBW), from \cite{Boulle:2007b} - Section~3.5}
\begin{itemize}
\item Multi-start: repeat $\log(KN)$ times:
	\begin{itemize}	
	\item Start with an empty subset of variables
	\item Fast Forward Backward Selection:
		\begin{itemize}
		\item Initialize an empty subset of variables
		\item Repeat until no improvement, at most two times:
			\begin{itemize}
			\item Randomly reorder the variables
			\item Fast Forward Selection
			\item Randomly reorder the variables
			\item Fast Backward Selection
			\end{itemize}
		 \end{itemize}
	\item Update the best subset of variables if improved
	\end{itemize}
\item Return the best subset of variables
\end{itemize}
\label{AlgoSNB}
\end{algorithm}

\section{Direct optimization of continuous weights for Naïve Bayes model}
\label{Direct}

We propose here a direct approach to estimate sparse and robust classifiers with continuous weights in $[0,1]^K$. This approach is based on the choice of two elements:
\begin{itemize}
\item{The choice of an evaluation criterion on the space of classifiers with continuous weights}
\item{The choice of an algorithm to optimize this criterion on the model space}
\end{itemize}

\subsection{The choice of a sparse regularized criterion}
\label{DirectRegularization}

The classification models we are interested in are of the \emph{weighted naive Bayes classifier} family. A given classification model denoted $M_w$ is described by its weight vector $w =(w_1,w_2,\ldots,w_K)$ in $\mathcal{B}  \overset{\textrm{def}}{=} [0,1]^K$.
As for the Selective Naive Bayes predictor, the fit of the data given the model can be evaluated using the negative log-likelihood of the data, which is given by: 
\begin{equation}
F_N(w) \overset{\textrm{def}}{=} - \sum_{n=1}^N LL_n(w), w \in \mathcal{B}  \overset{\textrm{def}}{=} [0,1]^K.
\label{Fns}
\end{equation}

We obtain exactly the same Formula~\ref{LLn}, except that the weights are continuous values in $[0,1]$ instead of Boolean values in $\{0,1\}$.
As in the Boolean case, the minimization of the negative log likelihood leads to an over-fitting problem, which is addressed by adding a regularization term that expresses constraints on the weight vector $w$.
The regularized criterion is: 
\begin{equation}
CR(w)= - \sum_{n=1}^N LL_n(w) + \lambda f_C(w),
\end{equation}
where  $LL_n$ refers to the log-likelihood, $f_C$ is the regularization function for continuous weights, and $\lambda$ is a positive regularization weight.\\

Several objectives have guided our choice of the regularization function:\\
\begin{itemize}
\item[1.]Its sparsity, i.e. it favors the weight vectors composed of as much null components as possible. The $L^p$ norm functions are usually employed in regularization terms of the form $\sum_{k=1}^K |w_k |^p$. These functions are increasing and hence favor the weight vectors with low components. 
For $p>=1$, the norm function $L^p$ is convex, which makes the optimization easier and renders this function attractive. This explain the success of $L^2$ regularization in many contexts. For ill-posed linear problems, the ridge regression also called Tikhonov regularization \cite{Hoerl:1970} uses the $L^2$ norm.
However, the minimization of the regularization terms for $p>1$ does not necessarily lead to variables elimination whereas the choice $p \leq 1$ favors sparse weight vectors.
The Lasso method and its variants \cite{Hastie:2015} exploit the advantages of the value $p=1$, which enables sparsity and convex optimization.
For $p<1$, the $L^p$ regularization better exploits the sparsity effect of the norm but conducts to non convex optimization. Alternatively, function $w^p$ with $p<1$ can be replaced by any function $\xi(w)$ that satisfies the following conditions $(C1)$ : $\xi(w)$ is increasing and concave on $[0,1]$, with $\xi(0)=0$ and $\xi(1)=1$.\\
\item[2.] Its derivability. Some functions satisfying the above conditions $(C1)$ have infinite derivative at 0. Therefore, in order to avoid numerical instability, we modify it in a small neighborhood $[0, \delta]$ of the origin. Namely, we introduce the function $$\xi_{\delta}(\tau) = \left\lbrace\begin{array}{cl}
\xi(\tau)& \tau \geq \delta \\
\xi(\delta) + \xi'(\delta) (\tau - \delta) & 0 \leq \tau \leq \delta \end{array}\right.$$ where $\delta$ is a small positive number. Finally, we replace function $\xi$ by the function $\frac{\xi_{\delta}(\tau) - \xi_{\delta}(0)}{1- \xi_{\delta}(0)}$. Note that this is mainly a "technical trick" motivated only by solving numerical optimization problems.\\
\item[3.]Its ability to take into account the $B(X_k)$ costs per input variable such that, for equivalent likelihoods, the "simple" variables are preferred to "complex" ones. By weighting the terms under the $L^p$ norm with these costs, we obtain a penalization criterion of the form: $\sum_{k=1}^K B(X_k) * |w_k|^p$. These variable costs are supposed to be known before optimization. If no knowledge is available, these costs are fixed to $1$, otherwise, they can be used to include expert knowledge. In our case, as in Section~\ref{AveragingCriterion}, these costs come from the preparation cost of the variable, i.e. the discretization cost for a numerical variable, resp. the grouping cost for a categorical variable, the selection and the construction costs (see \cite{BoulleEtAlML18} for details).\\
\item[4.]Its consistency with the regularized criterion of the MODL naïve Bayes classifier in the case of Boolean selection of variables \cite{Boulle:2007b} presented in previous section. In order to get a criterion $f_C$ almost identical to the criterion $f_B$ when $w_k$ are Boolean values, we use in $f_B$ the approximation $\log K_s ! = K_s (\log K_s -1) + O(\log K_s)$.
\begin{footnotesize}
\begin{eqnarray*}
\label{SNBPriorApprox}
f_B(w) &=&L^*(K_s) - \log {(K_s !)} + \sum_{K_s} B(X_k), \\
			 &=&\sum_{K_s} (B(X_k) - \log K_s + 1) + O(\log K_s).
\end{eqnarray*}
\end{footnotesize}
Altogether, this finally results in the following regularization term:
\begin{equation}
f_{C}(w)=\sum_{k=1}^K (1 - \log{\left( \langle \bar{e},w \rangle +1 \right)} + B(X_k)) \xi_{\delta}(w_k),
\end{equation}
where $\bar{e}$ is the vector of all ones. The compact notation $\langle \bar{e},w \rangle$ therefore refers to $\sum_{k=1}^K w_k$.
Let us precise that the $B(X_k)$ costs are large enough to ensure the condition  $1 - \log{\left( \langle \bar{e},w \rangle +1 \right)} + B(X_k) > 0$ for all feasible $w \in \mathcal B$.
\end{itemize}

\smallskip
While other criteria could bring sparsity, our choice was mostly guided by the last objective of being consistent with the regularized criterion in the case of binary selection of variables. As a matter of fact, this criterion was obtained from a Bayesian model selection approach, with solid theoretical foundation. We expect that the chosen criterion presented above will keep its appealing statistical properties while promoting sparsity.

\subsection{The minimization of the chosen criterion with non convex optimization algorithms} 
\label{optimization}

The work presented in this section is the result of a fruitful collaboration : we are very grateful to Yurii Nesterov who helped us to analyze our criterion and who proposed optimization algorithms adapted to our problem.
\smallskip

The problem is to minimize the objective function $CR(w)$ subject to the constraint that $w$ takes its values in $[0,1]^K$, i.e. to solve the optimization problem:
\begin{equation}
\min_{ w \in \mathcal B} \left[ CR(w) \overset{\textrm{def}}{=} F_N(w) + \lambda f_{C}(w) \right].
\label{globalpb}
\end{equation}
The objective function $CR(w)$ consists of two terms, the log-likelihood term and the penalization term.\\
The log-likelihood term is a convex function of $w$. To show this, let $\alpha_n = - \log{P(Y=y^{n})}$, $\beta_j = - \log P(Y=C_j)$, $a_n$ a vector with components $a_n^{(k)}=-\log{p(x_k^{n}|y^{n})}$ and $b_{n,j}$ vectors with components  $b_{n,j}^{(k)}= - \log{p(x_k^{n}|C_j)}$ for $k=1,\ldots,K$ be all constant quantities in this optimization problem.
Using a vectorial notation, the negative log-likelihood per instance can be written :
\begin{equation}
LL_n(w) =  \alpha_n - \langle a_n, w \rangle + \log{\left( \sum_{j=1}^J e^{- \langle b_{n,j}, w \rangle - \beta_j} \right)}
\label{LLnVect}
\end{equation}
The first and second term of $LL_n$ are resp. constant and linear in $w$ and then are both convex. The third term is convex because of the log-convexity of $\exp(x)$. The log-likelihood term is then a convex function of $w$. Its second derivatives are bounded.\\ 
The regularization term is more complicated: it is concave (use of the $L^p$ norm with $p < 1$) and its partial derivative are unbounded at the points with zero components (see derivability issue in Section~\ref{DirectRegularization}). This makes impossible to establish theoretical guarantees even for convergence to a local solution. 

To solve this optimization problem, we have compared algorithms which can be classified according to two main strategies, one-stage and two-stages strategies, described in the following sub-sections. The theoretical justification and details of all methods below can be found in monograph \cite{Nesterov:2004}  and papers \cite{Nesterov:2005,Nesterov:2013,Nesterov:2018}.

\smallskip
Let us note that although the compared algorithms are state of the art, they do not provide guarantees w.r.t. the gap between the obtained solution and the optimal one.
Still, our goal is obtain a well optimized solution, not necessarily the optimal one, in order to obtain parsimonious accurate classification models with scalabilitity and optimization time constraints.
Given this objective, strict optimality does not make sense, since our primary issue is a statistical problem of model selection from a data sample, with intrinsic variability. 

\subsubsection{One-stage strategies}
These strategies directly solve the minimization problem (\ref{globalpb}) by generating a sequence of vectors $w^t  \in  \mathcal B$ with monotonically decreasing value of the objective function $CR(w)$. If for some vector $w_t$ its $k^\mathrm{th}$ component becomes zero, then it will be fixed to zero for all next points. We implemented two local minimization strategies. 
\begin{itemize}
\item[(a)] \textit{Simplest Gradient descent (Nonconvex Optimization)} (SG). At each iteration of this scheme, starting from an input point $w_t$, and the scaling parameter $L=L_t$, we compute so-called \textit{Gradient Mapping} with respect to Euclidean norm. If its satisfies some acceptance condition, we use it as $w^{t+1}$; if not, we increase $L$ and compute new candidate points again. The stopping criterion in this method is related to the small norm of the displacement.
\item[(b)] \textit{Alternating Nonconvex optimization} (AM). Note that non convexity of $f_{\xi}(w)$ is related to interaction of the linear term ${\langle \bar{e},w \rangle}$ and the function $\xi(w)$. If we treat them separately, then the corresponding auxiliary problems can be efficiently solved. This leads to idea of alternating minimization.
\end{itemize}

\subsubsection{Two-stages strategies}
In order to improve the stability of the results, we propose to use two-stage methods. They work as follows.

\paragraph{\underline{\textit{A. First stage.}}} At this stage, we are trying to find an approximate solution of a \emph{convex relaxation} of our problem (\ref{globalpb}). It looks as follows:
\begin{equation}
\min_{w \in \mathcal B} \widetilde{CR} \overset{\textrm{def}}{=} F_N(w) + \lambda \sum_{k=1}^K (1 - \log(W_0 +1) + B(X_k)) w_k
\label{convexpb}
\end{equation}
where $W_0$ is an a priori estimate for $W = \sum_{k=1}^K w_k$.
In order to solve this problem, we can use one of the standard gradient methods for convex optimization. As the objective function has Lipschitz-continuous gradients, Fast Gradient Methods can also be used. These methods can be found in \cite{Nesterov:2004}. We have compared the following three methods:
\begin{itemize}
\item[(a)] Simplest Gradient method for Convex Case (SG)
\item[(b)] Universal Gradient method (UG)
\item[(c)] Conditional Gradient method (CG)
\end{itemize} 
Let $\tilde{w}$ be the output of the first stage and $\widetilde{W}$ the sum of its components.

\paragraph{\underline{\textit{B. Second stage.}}} The output of the first stage is used at the second stage in the following way.
The estimate $\widetilde{W}$ is used for defining the new objective function:
\begin{equation}
\widehat{CR}(w) \overset{\textrm{def}}{=} F_N(w) + \lambda  \sum_{k=1}^K (1 - \log(\widetilde{W} +1) + B(X_k)) \xi_{\delta}(w_k)
\label{secondpb}
\end{equation}
The vector $\tilde{w}$ is used as a starting point for the local optimization methods solving the non convex optimization problem $\min_{w \in \mathcal B}  \widehat{CR}(w)$. At this stage we used two local methods:
\begin{itemize}
\item[(a)] Gradient method with Upper Estimator (UE)
\item[(b)] Gradient method for Composite Function (CF): as the second term of the criterion is separable in $w$, it can be used, despite its convexity, in the framework of Composite Minimization \cite{Nesterov:2013}.
\end{itemize}

\medskip
Considering both one-stage and two-stages strategies, we compared a total of eight algorithms denoted by SG and AM for one-stage strategies, SG.CF, SG.UE, UG.CF, UG.UE, CG.CF and CG.UE for two-stage strategies, for all combinations between first and second stages. The next section presents the experiments carried out to compare these algorithms.

\section{Experiments}
\label{Exp}

In this section, we present the experimental setup and compare the optimization strategies presented in Section~\ref{Direct}. Then, we study the parameterization of the objective function and its impact on predictor performance, and focus on the impact of the initialization. Finally, we introduce an alternative optimization algorithm and compare its performance to those of the SNB predictor.

\subsection{Experimental setup}

\begin{table}
\begin{small}
\begin{center}
\setlength\tabcolsep{3pt}
\begin{tabular}{|l|c|c|c||l|c|c|c|}
\hline
Dataset & Ni  & Nv/NvInf & Nc  & Dataset & Ni  & Nv/ NvInf  & Nc  \\
\hline
Abalone & 4177 & 8/8& 28& MfeatPix & 2000 & 240/240& 10\\
Adult & 48842 & 15/14& 2&MfeatZer & 2000 & 47/47& 10\\ 
Australian & 690 & 14 /10.6&2&MovieLens & 1000209 & 3/3 & 5\\
Breast & 699 & 10/9& 2& Mushroom & 8416 & 22/21&2 \\
Bupa & 345 & 6/0.3& 2& PenDigits & 10992 & 16/16& 10 \\
CensusIncome &299285& 41/41&2&Phoneme & 2254 & 256/256 & 5 \\
Connect4 & 67557 & 42/40.1& 3& Pima & 768 & 8/6.7&2 \\
Crx & 690 & 15/11.5& 2 & PokerHand & 25010 & 10/0& 10\\
Digits & 60000 & 784/639 & 10&Satimage & 768 & 36/36 &6 \\
Flag & 194 & 29/10.8 &8& Segmentation & 2310 & 19/18&7  \\
ForestType & 581012 & 54/53.7 & 7&  Shuttle & 58000 & 9/9 &7 \\
German & 1000&24/5.5&2& SickEuthyroid & 3163 & 25/15&2 \\
Glass & 214 &10/8.8 & 6& Sonar & 208 & 60/11.8& 2 \\
Heart & 270&13/9.1& 2& Soybean & 376 & 35/35& 19 \\
Hepatitis &155 & 19/8&2&  Spam & 4307 & 57/55& 2 \\
Horsecolic & 368 & 27/10.1& 2&Svmguide1 & 7089 & 4/4 & 2 \\
Hypothyroid & 3163 & 25/9.1 &2&Svmguide3& 1284&22/18&2\\
Ionosphere & 351 & 34/33& 2& Thrombin & 1909& 139351/12229&2\\
Iris & 150 & 4/4&3& Thyroid & 7200 & 21/7& 3 \\
Isolet & 6238&617/617& 26 &  Tictactoe & 958 & 9/3.5 & 2 \\
LED & 1000 & 7/7& 10& Twonorm & 7400 & 20 -20& 2\\
LED17 & 10000 &24/7 &10& Vehicle & 846 & 18/18& 4 \\
Letter & 20000& 16/15 &26&  Waveform & 5000 & 21/19 &3 \\
MfeatFac & 2000 & 216/216& 10& WaveformNoise & 5000 & 40/19&3 \\
MfeatFou & 2000 & 76/76& 10&  Wine & 178 & 13/13& 3 \\
MfeatKar & 2000 & 64/64& 10&WDBC & 569 & 31/26& 2 \\
MfeatMor & 2000 & 6/6& 10& Yeast & 1484 & 9/7.5&10 \\
\hline
\end{tabular}
\caption{Description of the $54$ UCI datasets: Ni=instances number, Nv=initial number of variables, NvInf= number of informative variables, Nc=class number.}
\label{UCIData}
\end{center}
\end{small}
\end{table}

\begin{table}
\begin{small}
\begin{center}
\setlength\tabcolsep{4pt}
\begin{tabular}{|l|c|c|c||l|c|c|c|}
\hline
Dataset & Ni  & Nv/NvInf  & Nc  & Dataset & Ni  & Nv/NvInf  & Nc  \\
\hline
Christine	&	5418&1636/682.6&	2 &Alexis	&	54491	&	5000/635.8	&2\\
Jasmine	&	2984&144/53	&2&Dionis	&	416188	&60/54	&355\\
Madeline	&	3140	&	259/11.4&	2&Grigoris	&	45400	&	301561/1631.2&	2\\
Philippine	&	5832	&	308/115&	2&Jannis	&	54	&	83733/27	&4\\
Sylvine	&	5124	&	20/8.4&	2&Wallis	&	10000	&	193731/952&	11\\
\hline
Albert	&	425240	&78/36&	2&Dorothea	&	800	&	100000/82.2&	2\\
Dilbert	&	10000	&	2000/1682&	5&Newsgroups	&	13142	&	61188/5132.8&20\\
Fabert	&	8237	&	800/256.8&	7&Evita	&	20000	&	3000/348.8&2\\
Robert	&	10000	&	7200/7143.4&10&Helena	&	65196	&	27/27&	100\\
Volkert	&	58310	&	180/147	&10&Tania	&	157599	&	47236/7308&	2\\
\hline
\end{tabular}
\caption{Description of the $20$ AutoML Challenge datasets: Ni=instances number, Nv=initial number of variables, NvInf= number of informative variables, Nc=class number.}
\label{AutoMLData}
\end{center}
\end{small}
\end{table}

\begin{table}
\begin{small}
\begin{center}
\setlength\tabcolsep{4pt}
\begin{tabular}{|l|c|c|c|c|c|c|}
\hline
Dataset & Ni   & Nc  & NvInf(100) & NvInf(1000) & NvInf(10000)  \\
\hline
ImagesElephant &200 & 2 & 2.2 & 20.8 & 217.4 \\
Musk1 & 92 & 2 & 0 & 1.2 & 28.8\\
Musk2 &102 & 2 & 0.8 & 2.2 & 3.4 \\
MimlDesert &	2000 & 2 & 46.4 & 396.2 & 2887\\
MimlSea & 2000 & 2& 17.8 &64.4  & 298.8\\
MimlSunset &	2000 & 2 & 69 & 645.4 & 5342.8\\
MimlTrees & 2000 &	2 & 43.6 & 400.6 &  2840.6\\
JapaneseVowels &	640 &	9 & 99 & 988.6 & 9874\\
SpliceJunction &3178 & 3 & 29.4 & 304 & 2303.4\\
Diterpenes &	1503 & 23 &	79.4 & 657.8 & 1670.8\\
MutagenesisAtoms & 188 & 2 & 30.6 & 174.6 & 321.2 \\
Mutagenesis & 188 & 2 & 30.4 & 199.8 & 1307.4\\
CharacterTrajectory &	2858 & 20 &	92 & 938 & not tested\\
OptDigits & 5620 & 10 &	25.4 & 290.8 & 3977.4\\
AIDSMolecule &42687 &	2 & 53 & 465.4 & 3834.4 \\
MTConnect4	& 67557 & 3 & 20.2 & 298.8 & 5235.4 \\
MTDigits & 60000 &	10 & 35.6 & 372.2 & 4827.4\\
\hline

\end{tabular}
\caption{Description of the $17$ relational datasets: Ni=instances number, Nc=class number,NvInf= number of informative variables for $100$, $1000$ and $10000$ constructed variables.}
\label{MTData}
\end{center}
\end{small}
\end{table}

We conduct the experiments on three collections of datasets. $54$ datasets come from the repository at University of California at Irvine \cite{Dua:2017} and $20$ datasets from the AutoML Challenge \cite{Guyon:AutoML:2016} which are all tabular datasets. The third database collection is composed of $17$ relational datasets, most of them used in \cite{BoulleEtAlML18}. For these multi-table datasets, automatic variable construction was applied to build $100, 1000$ and $10000$ variables \footnote{except for the dataset CharacterTrajectory for which only $100$ and $1000$ variables have been used due to excessive optimization time}, leading to 50 tabular datasets. Tables \ref{UCIData}, \ref{AutoMLData} and \ref{MTData} give a summary of these datasets, which overall amount to 124 optimization problems.
All the experiments are performed using a stratified cross validation in five or ten folds according to the size of the related optimization problem.

In the rest of this section, the following criteria will be considered to compare the different algorithms:
\begin{itemize}
	\item optimized criterion (\ref{globalpb}),
	\item number of selected variables (with weight greater than zero),
	\item optimization time,
	\item predictor performance on a test dataset:
	\begin{itemize}
		\item Accuracy (ACC), 
		\item Area Under the ROC Curve (AUC),
		\item Compression rate (Comp), that is the ratio of the log likelihood of the model to that of null model (with no selected variable).
	\end{itemize}
\end{itemize}

\subsection{Comparison of the optimization strategies}

The objective of this first study is to compare the performance of the eight minimization algorithms described in Section \ref{optimization}. To measure their performance, we compare the final value of the optimized criterion (\ref{globalpb}) and the final number of variables with positive weight. The total execution time is also studied to compare algorithms with equivalent performance. As we focus on the optimization quality, the $\lambda$ regularization coefficient is fixed to a high value, i.e. to $1$, to compare the methods in challenging situations. The parameter $p$ is fixed to $0.95$. For all methods, the initial solution is $w^0=\frac{1}{2} \bar{e}$ i.e. a uniform weight equal to $0.5$ per variable.\\
\begin{figure}[!tbp]
\begin{center}
\includegraphics[width=0.65\textwidth, keepaspectratio]{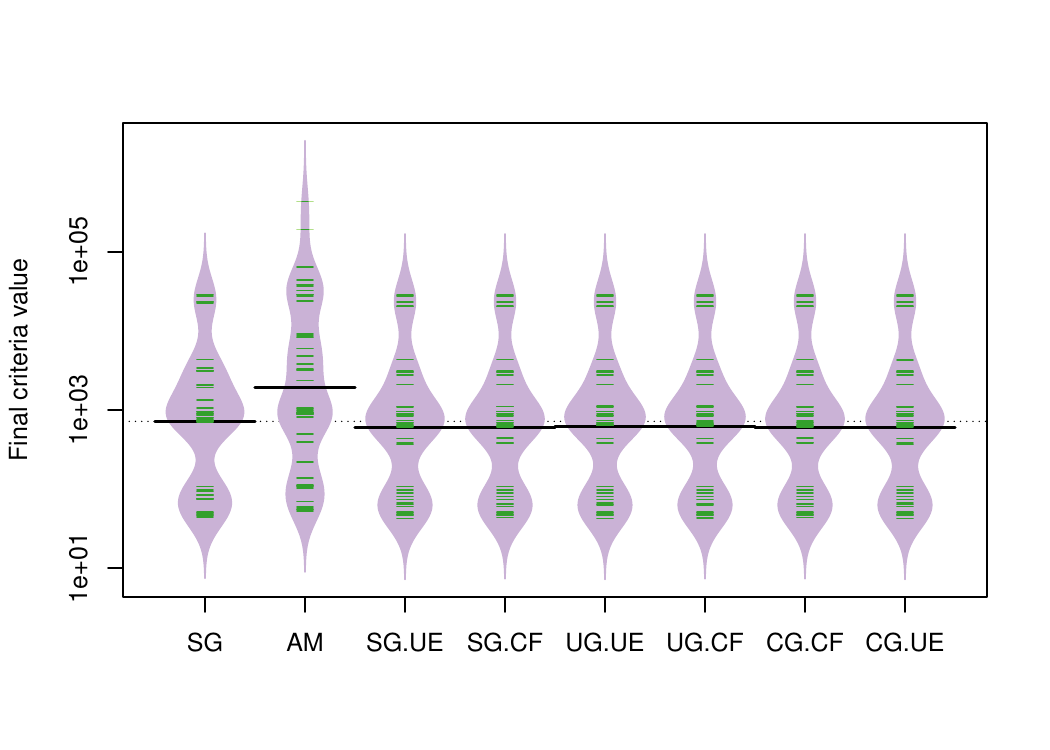}
\includegraphics[width=0.65\textwidth, keepaspectratio]{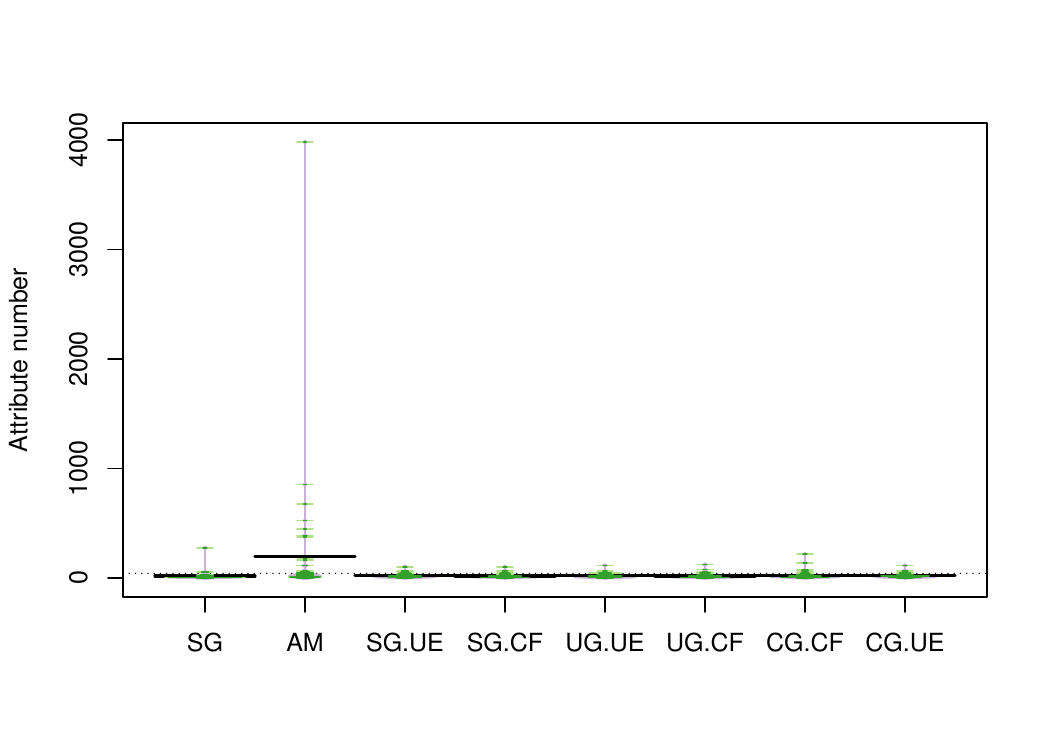}
\end{center}
\caption{Final criterion value and final variable number for the eight methods}
\label{8Algos-CritereParcimonie}
\end{figure}
\begin{figure}[!tbp]
\begin{center}
\includegraphics[width=0.65\textwidth, keepaspectratio]{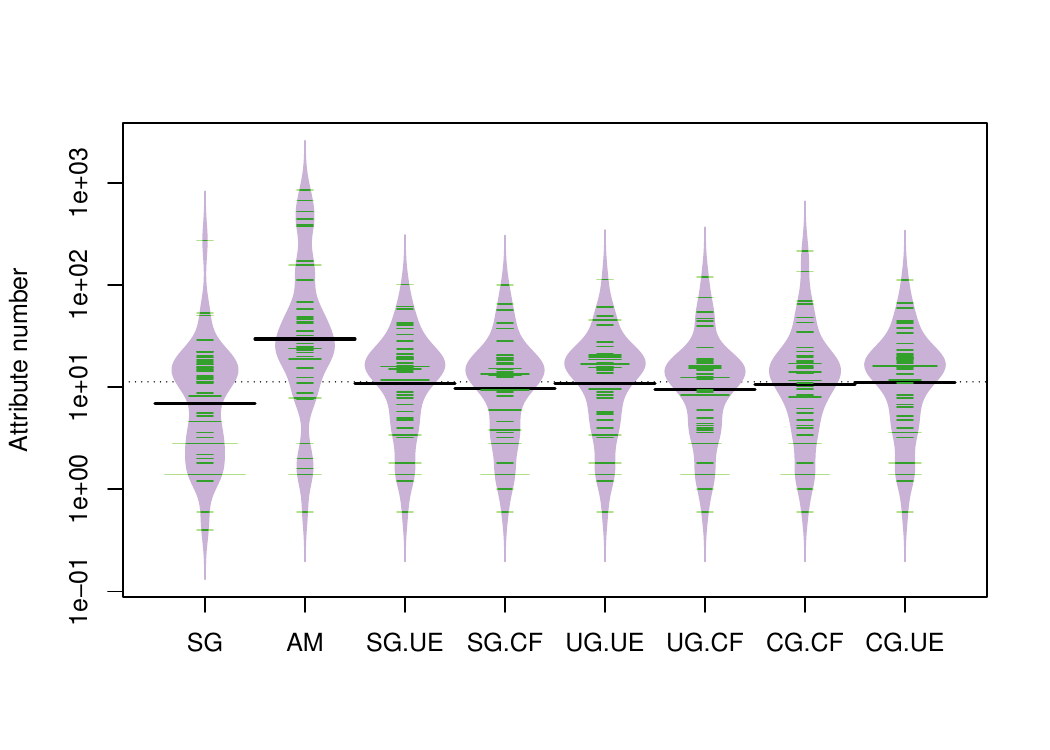}
\end{center}
\caption{Final variable number for the eight methods using a log scale}
\label{8Algos-CritereParcimonieSeuille}
\end{figure}
Figure \ref{8Algos-CritereParcimonie} presents the beanplots for the final criterion value and  the final number of variables for all the datasets. In these beanplots, each green horizontal line corresponds to the indicator value (final criterion or variable number) for one dataset. The bold black horizontal line represents the distribution mean. The bandwidth of the violet areas shows the smoothed density of the distribution. All these elements give a more complete view of the distributions than more summarized plots such as box whisker plots.\\
By looking at the final criterion value distributions on the top, we can first of all remark that these distributions are multi-modal : as the criterion is a sum over all the instances, it reflects a wide variety of the datasets in this benchmark, with varying numbers of instances or classes.\\
Secondly, the AM one-stage method appears to be the least effective: the AM method optimization quality is worse as its final criterion values are higher than those obtained with the other methods. The bottom plot of Figure \ref{8Algos-CritereParcimonie} is difficult to analyze as, for one dataset, the number of variables for the AM method is far higher than for the other datasets. Figure \ref{8Algos-CritereParcimonieSeuille} presents the beanplots obtained without this dataset i.e. \textit{JapaneseVowels} with $10000$ constructed variables. We can see that the models obtained with the AM method are less parsimonious as the related selected variable numbers are higher that those obtained with the other methods. \textbf{As a result, the AM method is excluded in the further result analysis}.

\begin{figure}[!tbp]
\begin{center}
\includegraphics[width=0.65\textwidth, keepaspectratio]{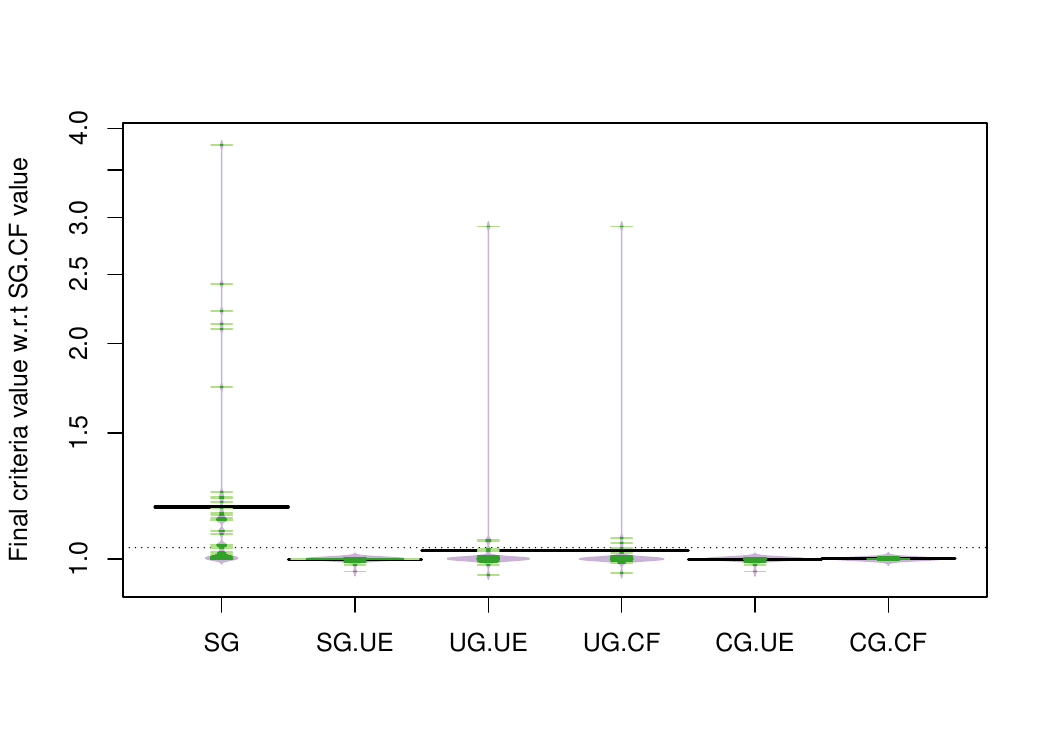}
\includegraphics[width=0.65\textwidth, keepaspectratio]{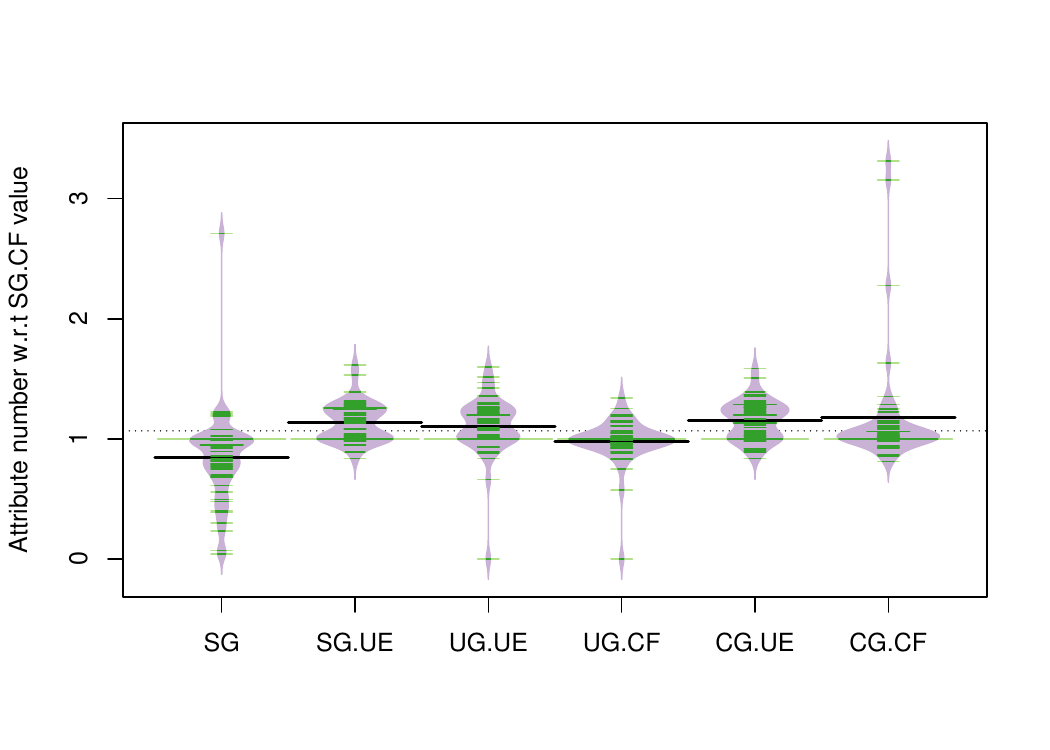}
\includegraphics[width=0.65\textwidth, keepaspectratio]{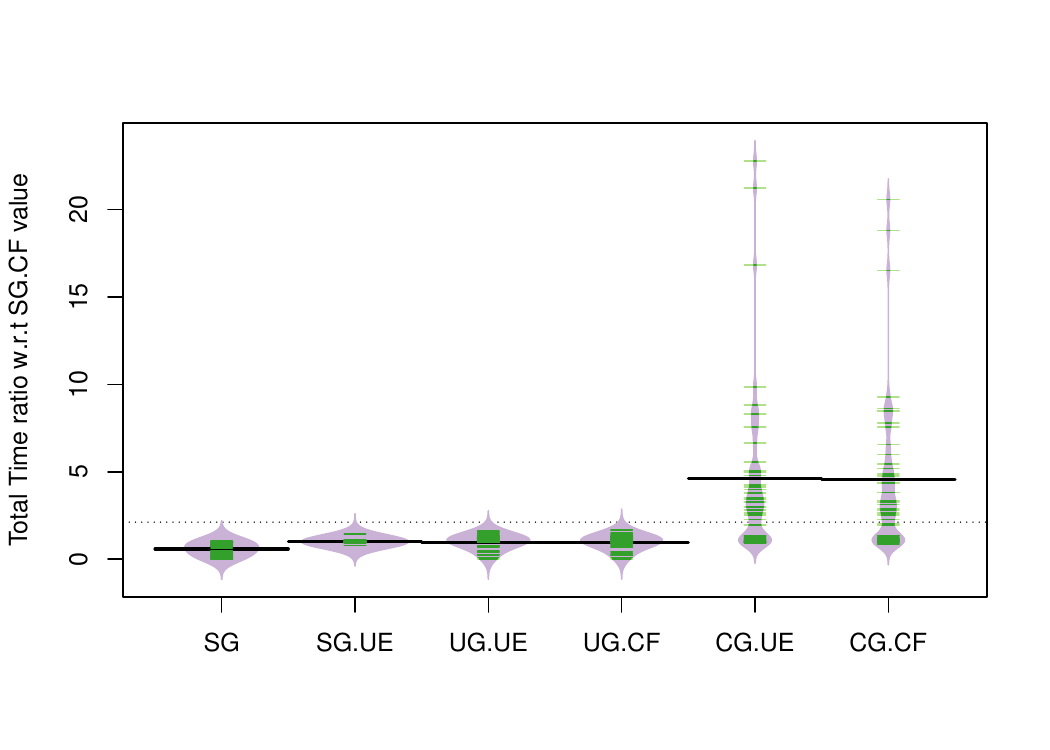}
\end{center}
\caption{Final criterion value, final variable number and execution time w.r.t their values for SG.CF method (excluding AM method)}
\label{7Algos-CritereParcimonie}
\end{figure}

Considering the other methods, the SG.CF method seems to be among of the best ones : that is why the ratios to the SG.CF indicators are presented in Figure \ref{7Algos-CritereParcimonie}.
The top graph shows the final criterion values obtained with the different algorithms compared to those obtained with the SG.CF method. On the left, the final criterion value ratios obtained with the SG method are always greater than $1$ and exceed $1.5$ for several datasets. For these datasets, \textit{Character}, \textit{Diterpenes} and \textit{JapaneseVowels}, the SG method converges towards the model with zero weights for all variables, called the null model. These datasets are among the most complex datasets because of a larger number of class values, suggesting that the SG method is less robust. \textbf{Therefore, we choose to set the SG method aside}.

Note that all the remaining methods are two-stages methods.
Concerning the CG.UE and the CG.CF on the right, the final criterion values are similar to those of SG.CF as the criterion ratios are close to $1$ for all datasets. Nevertheless, the variable number ratios and especially the execution time ratios are high, which shows that CG.UE and CG.CF methods are less parsimonious than SG.CF method and are much more expensive as far as execution time is concerned. 
\textbf{For this reason, the CG.UE and the CG.CF methods are no longer considered in the following}. 

Three methods are remaining in the comparison with the SG.CF method : the SG.UE, UG.UE and UG.CF methods. With regard to the final criterion values, the two-stages UG methods are significantly less efficient for one dataset, \textit{JapaneseVowels}, that has also been poorly optimized by the AM and SG methods. This lack of efficiency is confirmed by looking at the final variable number distribution. For the UG.UE and UG.CF methods, we observe in Figure~\ref{7Algos-CritereParcimonie} an abnormally low variable number ratio, equal to zero, for one dataset, again \textit{JapaneseVowels}. For this dataset, the UG methods directly converges towards the so-called null model \footnote{The null model is the model with all weights equal to zero} without any variables. This could be explained by a higher sensibility to the initialization vector : in this case the initial vector with all variable weights equal to $0.5$ results in a high criterion value, far above that of the null model criterion value, and the the UG method converges towards the null model within one single iteration. To validate this hypothesis, we have normalized the initial vector using the ratio of the criteria obtained either using the vector with all weights equal to $0.5$ or the the null model with all weights equal to 0. With this initialization normalization, we obtained comparable results for UG and SG methods for the dataset \textit{JapaneseVowels} dataset with a final criterion ratio equal to $0.998$. \textbf{Nevertheless, this lower robustness of the UG methods w.r.t initialization made us opt for the SG methods}.

For the two remaining methods, SG.CF and SG.UE, the final criterion values and execution times are equivalent. However, the SG.UE method is a bit less parsimonious than SG.CF, as the ratio of the final variable number is almost always greater than one. \textbf{At last, we adopt the SG.CF method for the next experimentation as the reference among the continuous optimization methods.}

\subsection{Parametrization of the SG.CF method}

\begin{figure}[!tbp]
\begin{center}
\includegraphics[width=0.65\textwidth, keepaspectratio]{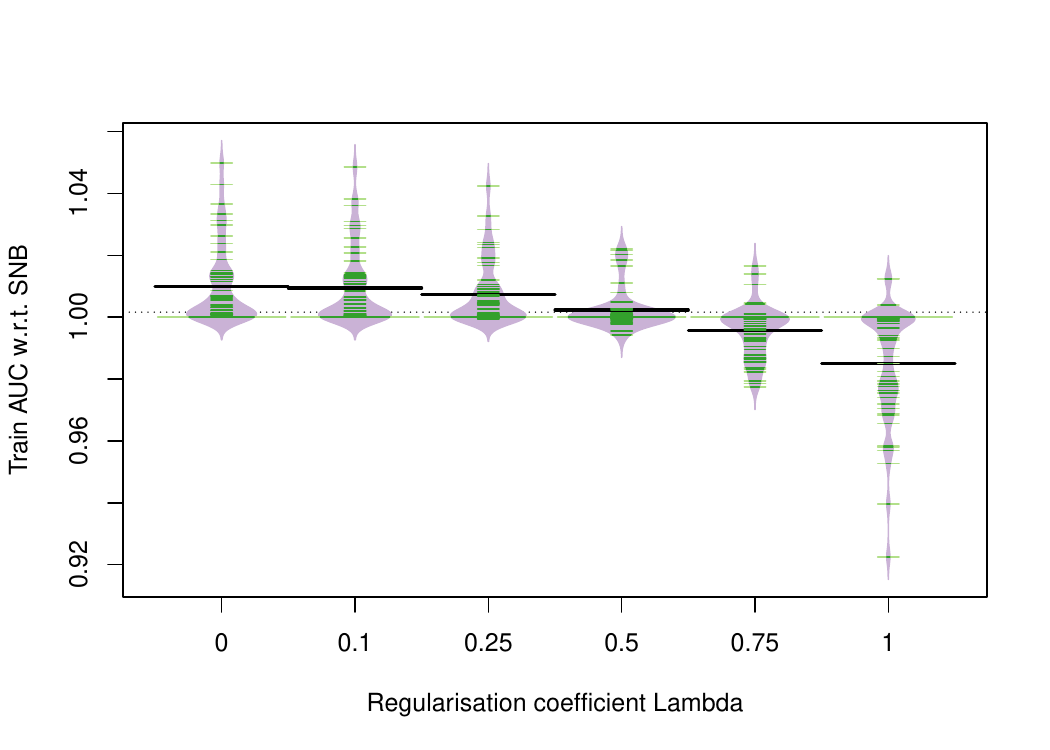}
\includegraphics[width=0.65\textwidth, keepaspectratio]{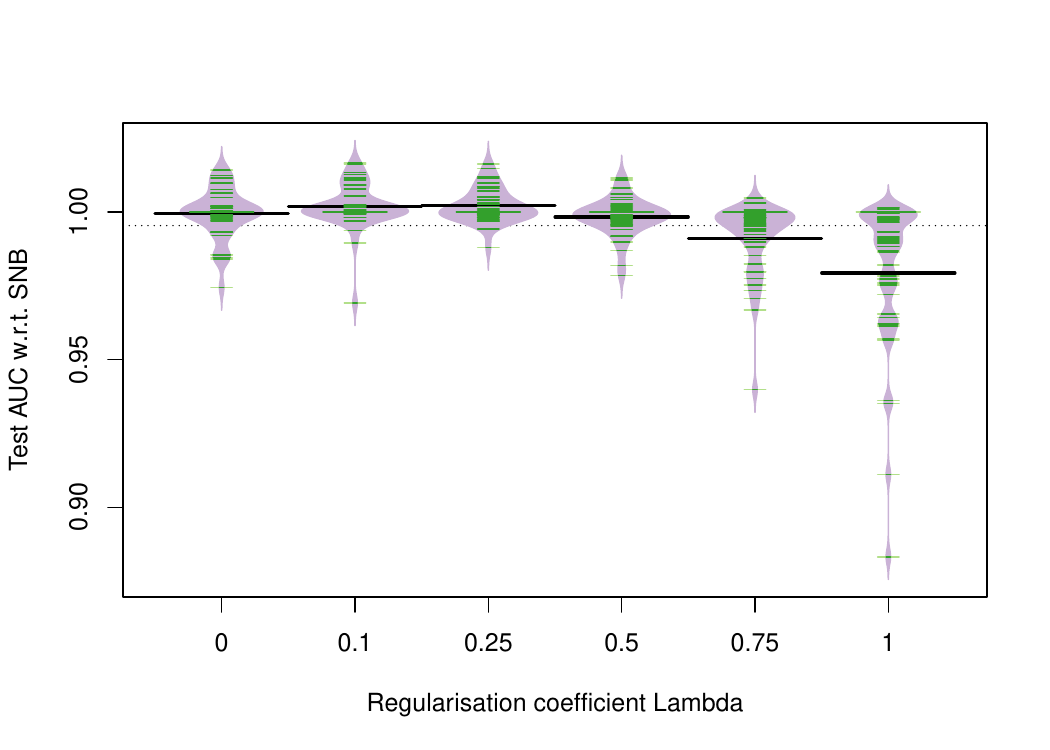}
\end{center}
\caption{Train and test AUC for SG.CF w.r.t SNB for different values of the regularization coefficient $\lambda$}
\label{ASLambda01}
\end{figure}

The objective of the following experimentation is to finalize the parameterization of the SG.CF method concerning the regularization coefficient $\lambda$, the regularization exponent $p$ and the $\epsilon$ value used for the stopping criterion. \\
For small $\lambda$ and in particular for $\lambda=0$, the absence of regularization conducts to over-fitting models with high AUC for train datasets and deteriorated AUC for test datasets. We are interested in high performance for test datasets rather than for train datasets. We expect to achieve this using higher values of the regularization parameter $\lambda$, leading to slightly worse train performance but better test performance. Figure \ref{ASLambda01} presents the train and test AUC of the SG.CF method w.r.t. to the AUC of the SNB model for $\lambda=0, 0.1, 0.25, 0.5, 0.75, 1$ for fixed $p=0.95$. To give an idea of the performance of the SG.CF with respect to the performance of SNB algorithm, the plots present the ratio of performance between the SG.CF  and SNB methods, both for the train and test datasets.
As expected, the train AUC decreases when $\lambda$ increases. Concerning the test AUC, the best ratios are obtained for $\lambda=0.25$ which stands out as the optimal value for this study.  The mean value for test AUC for $\lambda=0.25$ is of the order of $1$, meaning that the SG.CF method has similar prediction performance as the SNB method.

Having set $\lambda=0.25$, we have performed a sensitivity analysis of the regularization exponent $p$.
The cases of strictly convex ($L^2$), convex ($L^1$) or concave ($L^p, p < 1$) penalization terms were evaluated in \cite{HueEtAl2017} and showed a clear advantage of using a concave penalization term w.r.t. sparsity of the solution.
Moreover, the $L^1$ norm does not favor sparsity: in the case of two identical variables, any pair of variables weights $(w_1, w_2)$ with same total weight $w = w_1+w_2$ results in exactly the same criterion value.

\begin{figure}[!tbp]
\begin{center}
\includegraphics[width=0.65\textwidth, keepaspectratio]{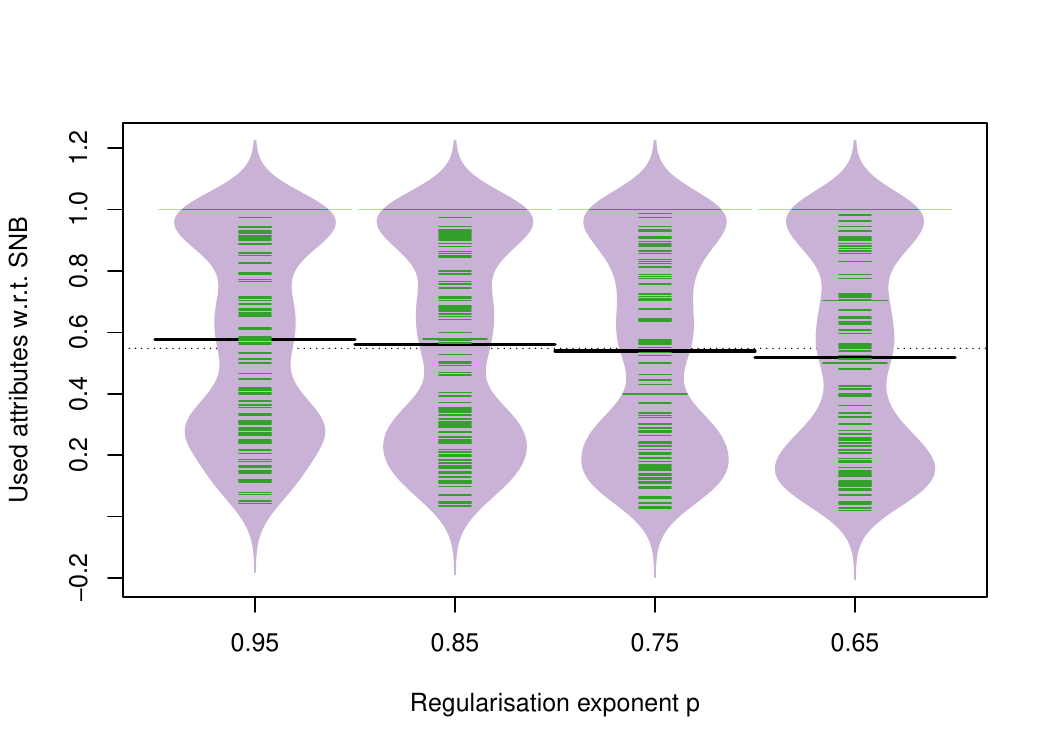}
\includegraphics[width=0.65\textwidth, keepaspectratio]{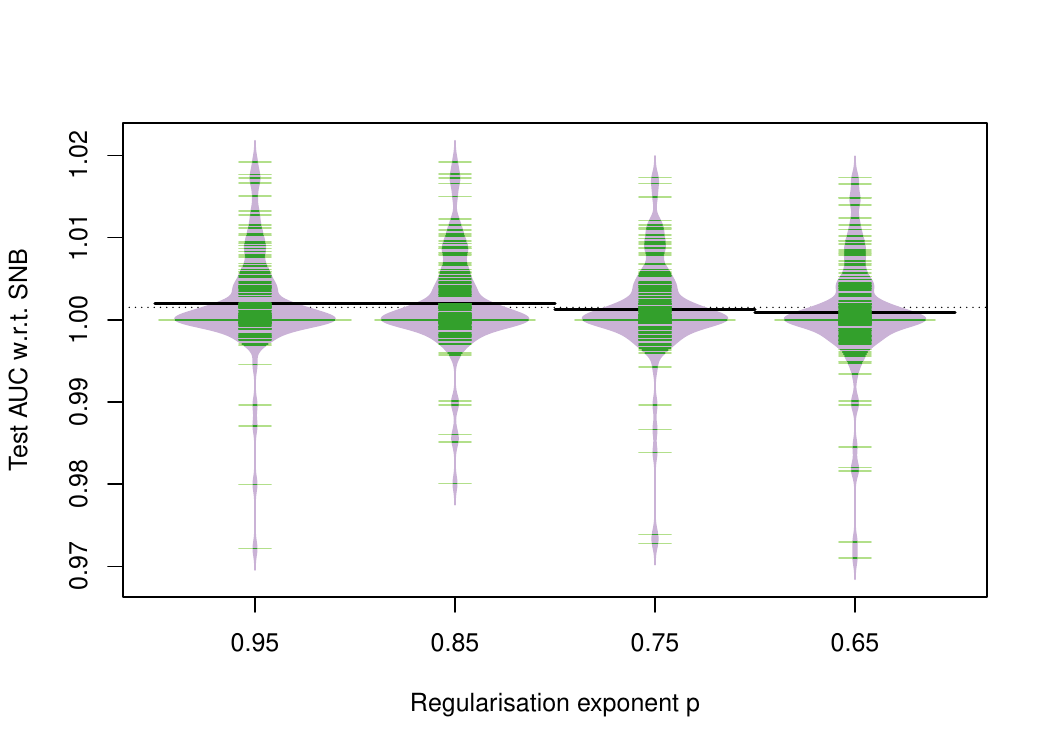}
\includegraphics[width=0.65\textwidth, keepaspectratio]{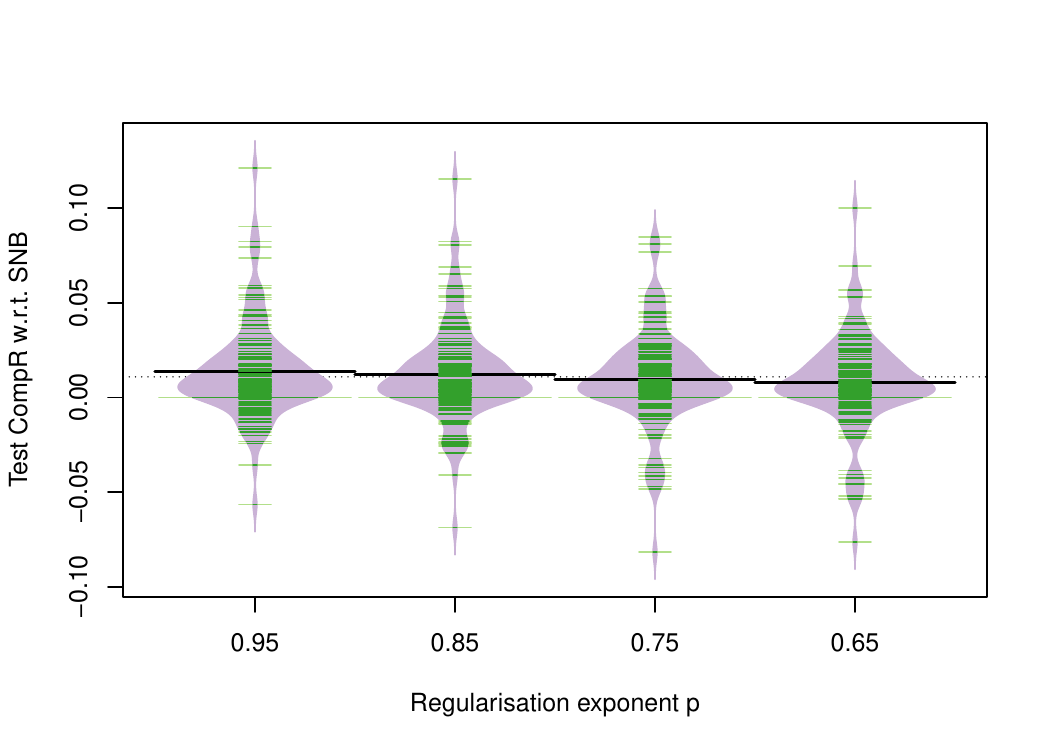}
\end{center}
\caption{Number of selected variables, test AUC and compression of SG.CF w.r.t SNB for different values of the regularization exponent $p$}
\label{AS-P}
\end{figure}

In this paper, we do not consider the case of the $L^1$ norm and evaluate the values $p=0.95, 0.85, 0.75, 0.65$.
As the penalization term increases when $p$ decreases, we expect that the smaller the $p$ value, the smaller the number of selected variables in the final model. This trend is well observed on top of Figure \ref{AS-P}, which presents the ratio of the number of variables of the SG.CF method to that of the SNB method. Concerning the predictive performance, we observe that the AUC on test datasets slightly decreases with $p$. As a matter of fact, as soon as $p$ is strictly smaller than $1$, the regularization term enables to reduce the variable number. Too small values of $p$ are not necessary and  may lead to performance degradation. That is why we choose to set $p$ to $0.95$.

\begin{figure}[!tbp]
\begin{center}
\includegraphics[width=0.65\textwidth, keepaspectratio]{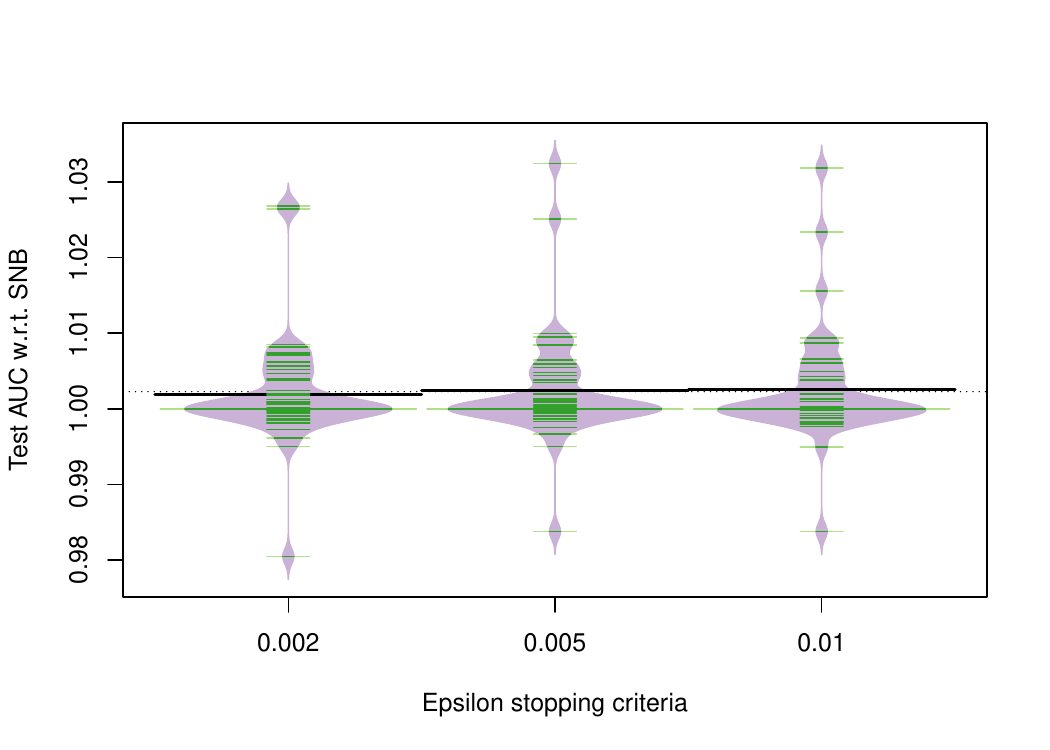}
\includegraphics[width=0.65\textwidth, keepaspectratio]{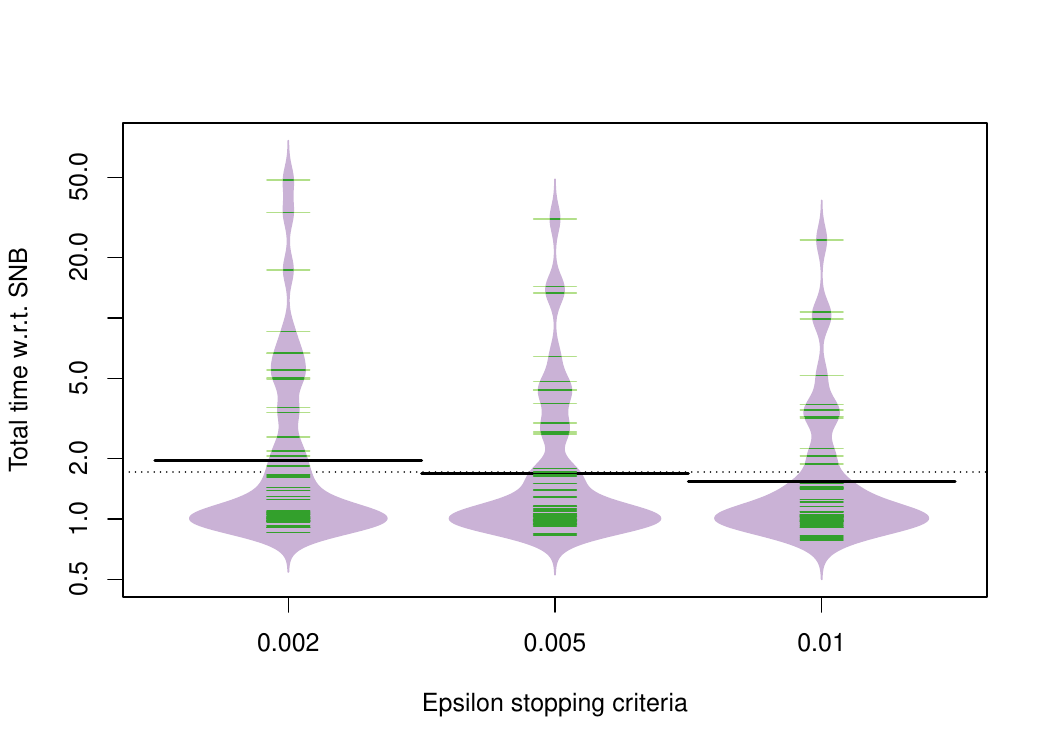}
\end{center}
\caption{Test AUC and optimization time for SG.CF w.r.t SNB for different values of the stopping criterion $\epsilon$}
\label{ASEps}
\end{figure}

A sensitivity analysis for the stopping parameter $\epsilon$ has also been conducted. The smaller $\epsilon$, the higher the iteration number and the better the minimization. Increasing the $\epsilon$ value enables to reduce the execution time at the expense of performance.
In Figure \ref{ASEps}, ratios of the test AUC and execution time with respect to the related SNB indicators are shown for $\epsilon=0.002, 0.05$ and $0.01$. There is no degradation of test AUC when $\epsilon$ is increasing. Execution time is decreasing as $\epsilon$ is increasing but this decrease is very slow. Even for $\epsilon=0.01$, execution time can be $10$ to $20$ larger for SG.CF than for SNB. These results have led us to parametrize the $\epsilon$ value to $0.01$ and to study how the execution time could be reduced with a better initialization of the SG.CF algorithm.

\subsection{Impact of the initialization}

In all previous experiments, the minimization algorithms have been initialized with a uniform weights vector $w^0=\frac{1}{2} \bar{e}$ i.e. a uniform weights equal to $0.5$ per variable. To reduce the execution time, we have tested two other initializations. The first one uses the weights obtained with the SNB algorithm described in Algorithm \ref{AlgoSNB}. 
An extension of the SNB algorithm has also been proposed to provide an initialization for the SG.CF algorithm with sparse fractional weights.
The space of the Selective Naïve Bayes classifiers with weights in $\{0,1\}^K$ is expanded to the space of the naïve Bayes classifiers with fractional weights in $[0,1]^K$. The same Fast Forward Backward algorithm is applied as described in Algorithm \ref{AlgoSNB} except that the "binary" adds or drops of variables are replaced by adds or drops using a fractional increment or decrement of weight. The fractional increment/decrement is chosen of the form $1/2^i$ and decreasing from $1/2$ as long as $1/2^i > 1 / N$. The pseudo-algorithm is described in Algorithm \ref{AlgoFNB}. The specificities of this FNB \footnote{Implemented and parallelized in the last version of the Khiops tool\cite{Khiops10} for classification and regression.} algorithm w.r.t. the SNB algorithm are presented in italics : we use only one start and the number of FFWBW passes for a given weight is fixed to $\log(KN)/\log(N)$, such that the overall algorithmic complexity is equal to $O(KN \log(KN))$, the same as that of the SNB algorithm. We can notice that the obtained FNB is defined by the best subset of fractional weights and is not obtained by averaging like for the SNB.\\

\begin{algorithm}[ht]
\caption{Fractional Naive Bayes algorithm (FNB)}
\begin{itemize}	
\item Start with an empty subset of variables
\item Fast Forward Backward Selection:
	\begin{itemize}
	\item Initialize an empty subset of variables
	\item For $i=1,\ldots,I$ times \textit{while weight increment $\frac{1}{2^i}<\frac{1}{N}$:}
		\begin{itemize}
		\item\textit{Repeat $1+\log K / \log N$ times :}
			\begin{itemize}
			\item Randomly reorder the variables
			\item Fast Forward Selection \textit{with weight increment equal to $\frac{1}{2^i}$}
			\item Randomly reorder the variables
			\item Fast Backward Selection \textit{with weight decrement equal to $\frac{1}{2^i}$}
			\end{itemize}
		 \end{itemize}
	\item Update the best subset of fractional weights if improved
	\end{itemize}
\item Return the best subset of fractional weights
\end{itemize}
\label{AlgoFNB}
\end{algorithm}

Each model with fractional weights is evaluated through a Bayesian criterion whose expression is given by:
\begin{equation}
\label{FNBCriteria}
- \sum_{n=1}^N LL_n(w) + \lambda  f_F(w)
\end{equation}
where
\begin{equation}
f_F(w) =  L^*( \lceil \sum_{K_s} w_k \rceil) - \log {(\lceil \sum_{K_s} w_k  \rceil !)} + \sum_{K_s} B(X_k) w_k^p
\end{equation}
Compared to the SNB criterion using $f_B(w)$ prior, the Boolean variable selection is replaced by a weighted variable selection with real values in $[0,1]$:
\begin{itemize}
\item the integer number of selected variables $K_s$ is replaced by the ceiling of the weights sum $\left \lceil \sum_{K_s} w_k \right \rceil$,
\item the cost $B(X_k)$ of each variable is multiplied by the continuous weight $w_k$ at the power $p$. The use of an exponent $p<1$ favors sparse models.
\end{itemize}
For Boolean weights, the FNB criterion exactly coincides with the SNB criterion, and for real value weight, the FNB criterion is very close with the WNB criterion (given the Stirling approximation $\log N! \approx N(\log N -1)$).

\begin{figure}[!tbp]
\begin{center}
\includegraphics[width=0.65\textwidth, keepaspectratio]{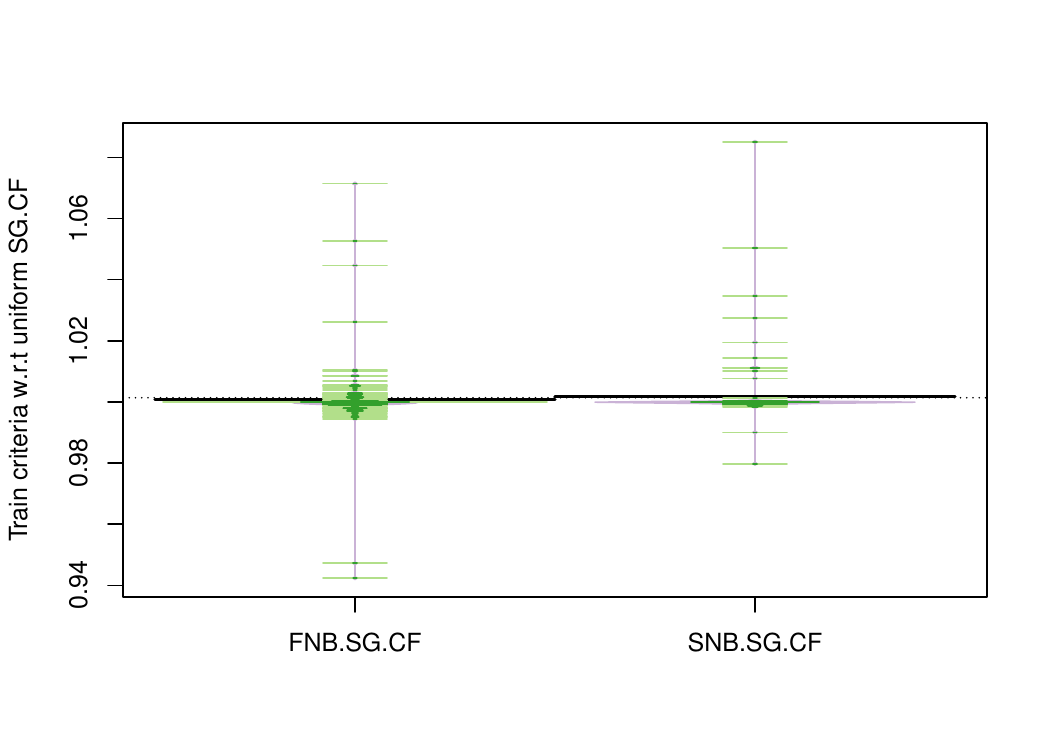}
\includegraphics[width=0.65\textwidth, keepaspectratio]{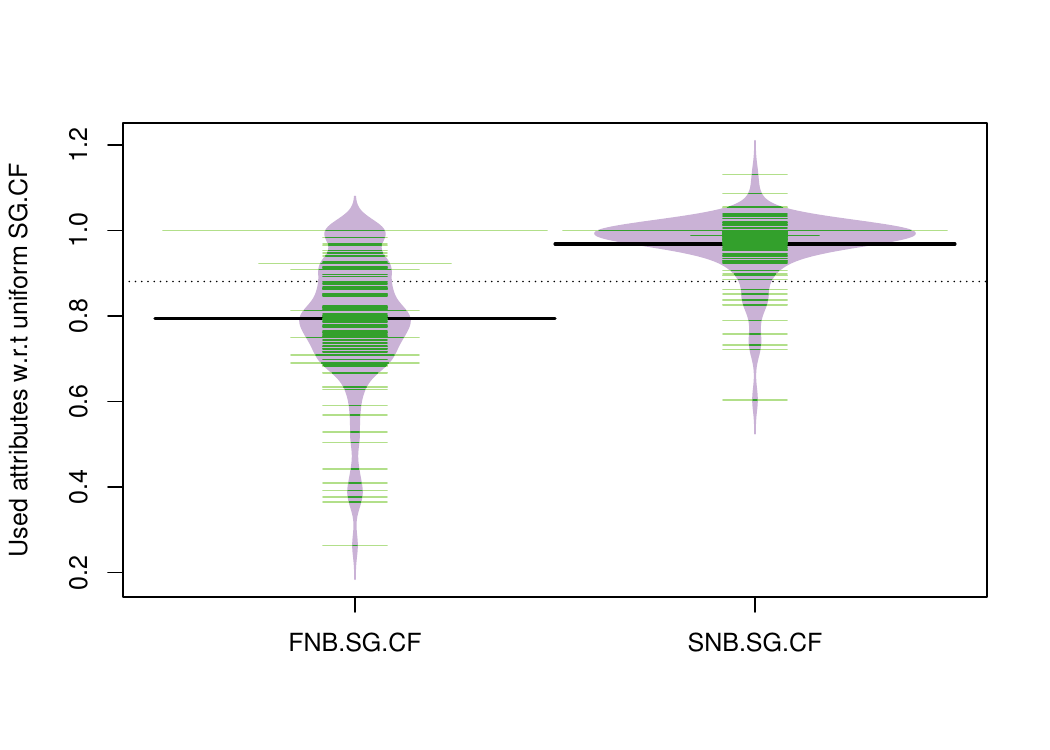}
\includegraphics[width=0.65\textwidth, keepaspectratio]{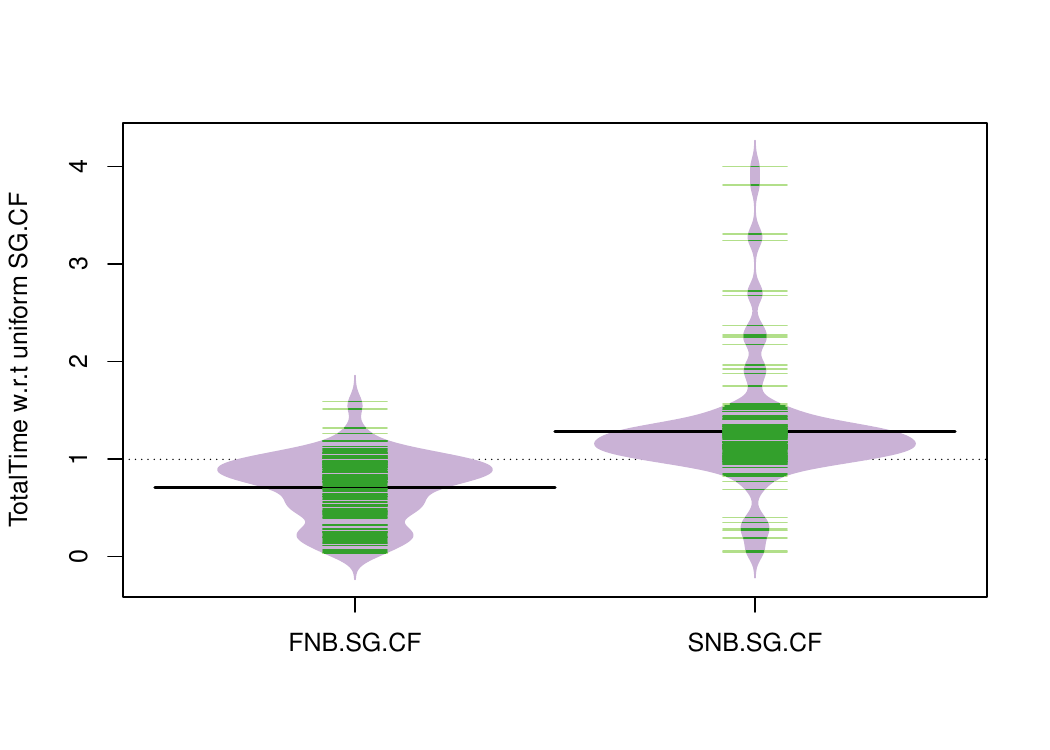}
\end{center}
\caption{Optimization performance of the SG.CF algorithm according to the initialization}
\label{Initialization}
\end{figure}

\begin{figure}[!tbp]
\begin{center}
\includegraphics[width=0.65\textwidth, keepaspectratio]{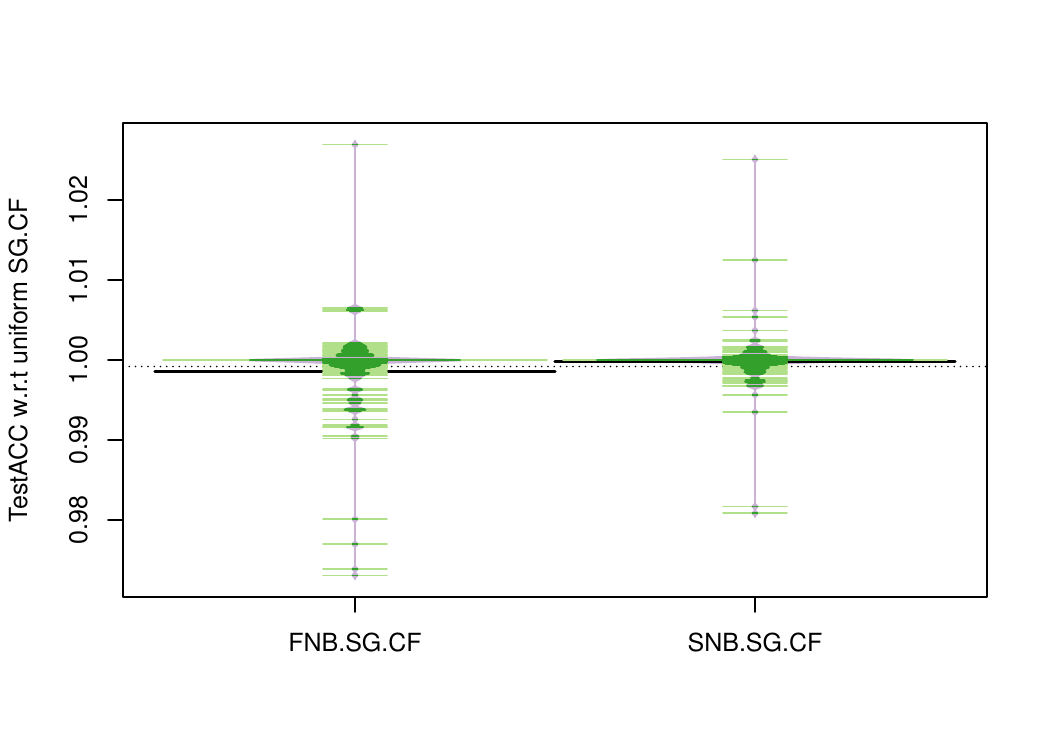}
\includegraphics[width=0.65\textwidth, keepaspectratio]{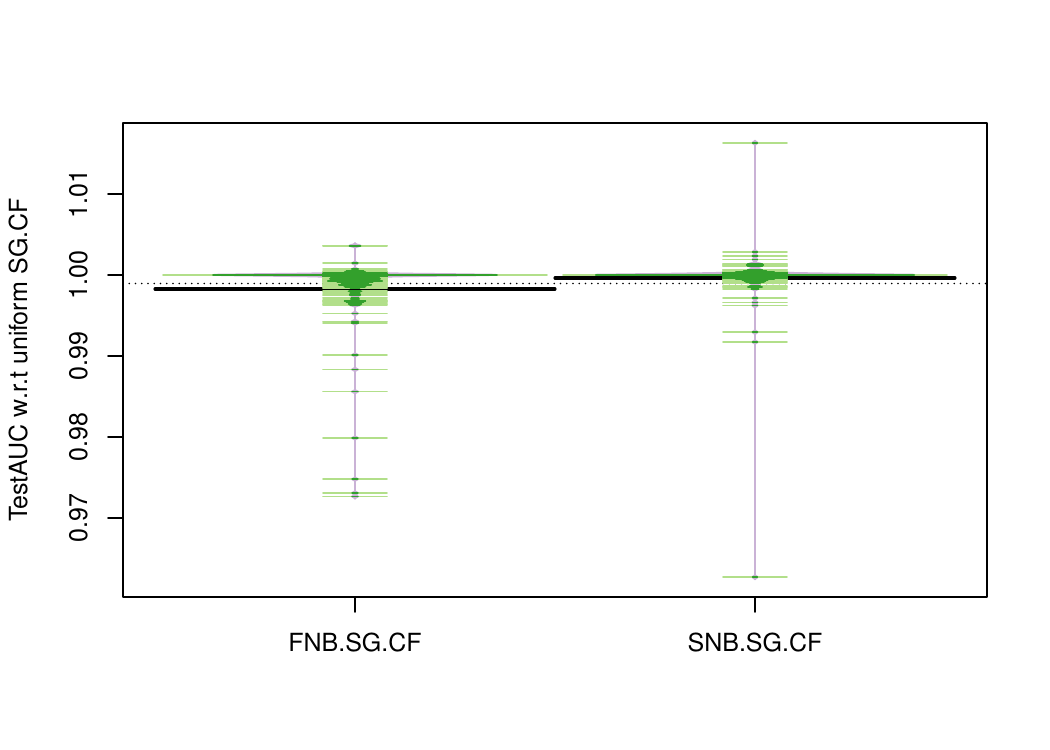}
\includegraphics[width=0.65\textwidth, keepaspectratio]{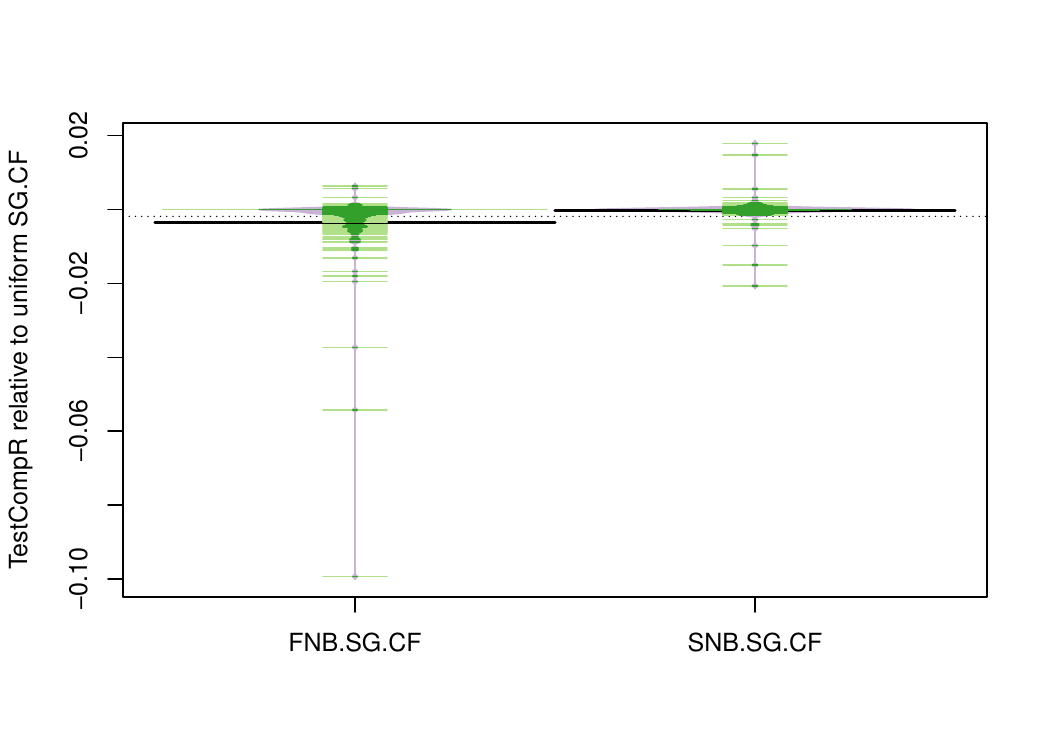}
\end{center}
\caption{Predictive performance of the SG.CF algorithm according to the initialization}
\label{InitializationPred}
\end{figure}

We have compared the performance of the SG.CF method according to the following initializations: uniform, SNB or FNB. 
Figure~\ref{Initialization} shows the impact of the initialization on the quality of the optimization : it presents the beanplots of the ratios of the train criterion, the number of selected variables and the execution time for the FNB and SNB initializations w.r.t. uniform initialization.  For most datasets, the final train criterion ratio is between $0.98$ and $1.02$ for the FNB and SNB initializations. The ratio of the number of selected variables is around $1.0$ for the SNB initialization and around $0.8$ for the FNB initialization. Concerning the execution time, the SNB initialization leads to a similar time but the FNB initialization enables to reduce it by $20\%$. The predictive performance for these initializations are presented in Figure \ref{InitializationPred}. The test ACC and AUC slightly decrease for the FNB and SNB initializations but never more than $3\%$. 
At the end, the SNB initialization does not seem interesting as it provides similar performance as the uniform initialization, but does not come with a decrease of the optimization time. On the other hand, the FNB initialization appears to be the best trade-off between execution time, quality of the optimization, number of selected variables and predictive performance. Overall, the FNB algorithm speeds up the convergence of the minimization algorithm by providing an initial weight vector in an interesting area.

\subsection{A competitive Fractional Naive Bayes}

\begin{figure}[!tbp]
\begin{center}
\includegraphics[width=0.65\textwidth, keepaspectratio]{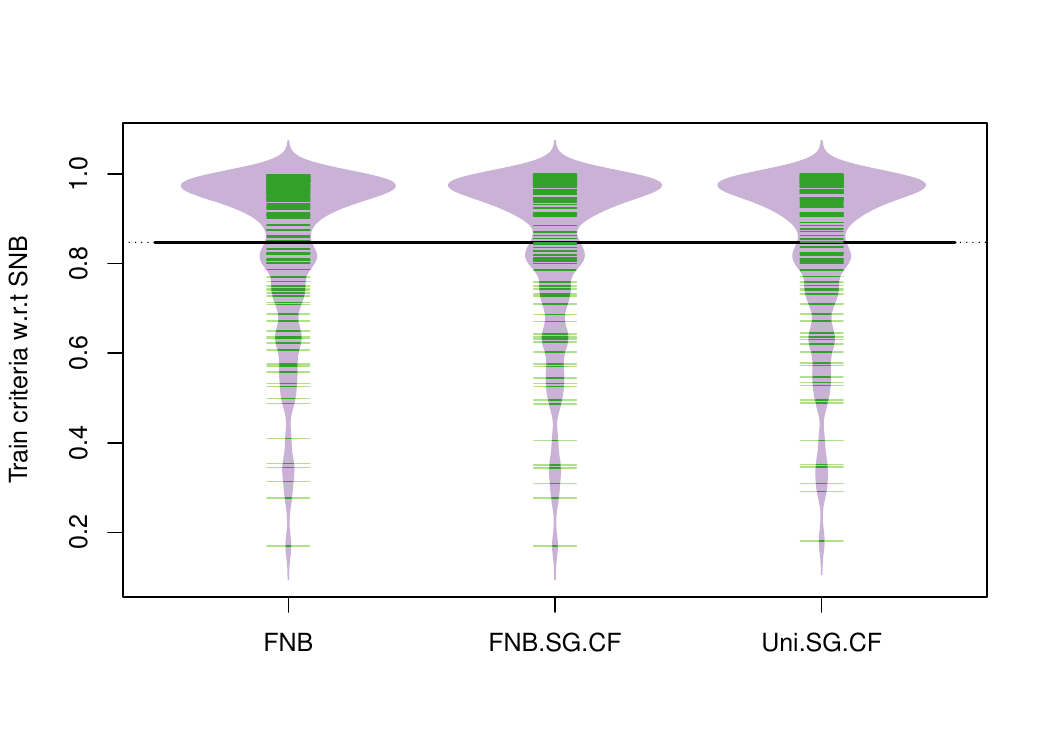}
\includegraphics[width=0.65\textwidth, keepaspectratio]{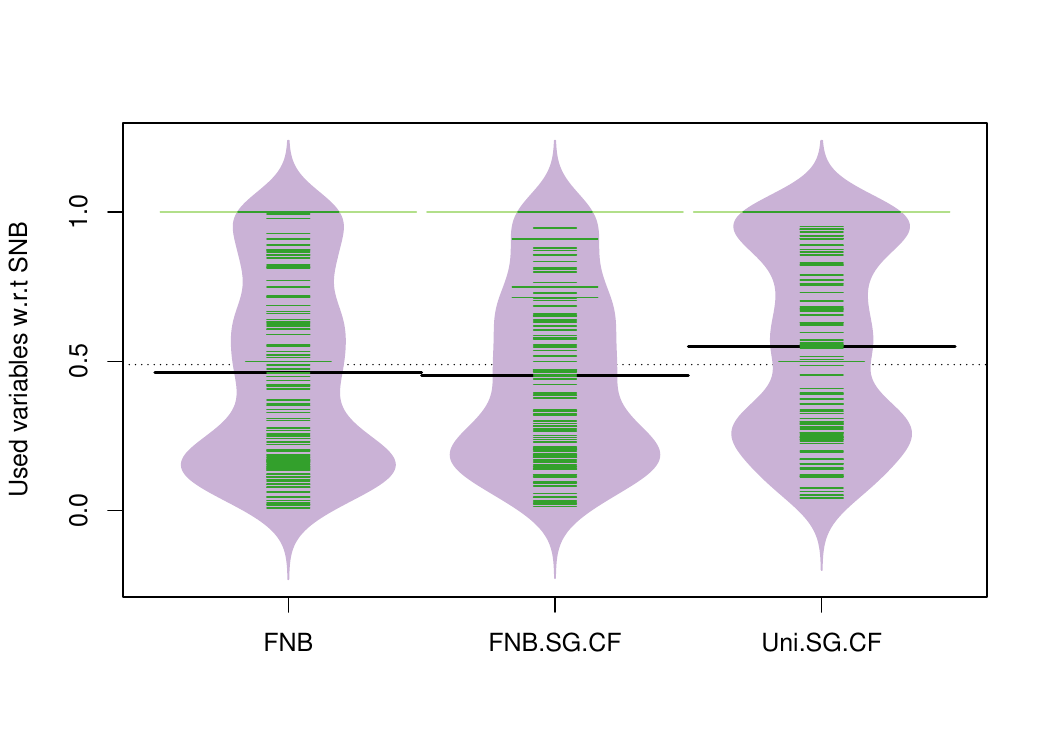}
\includegraphics[width=0.65\textwidth, keepaspectratio]{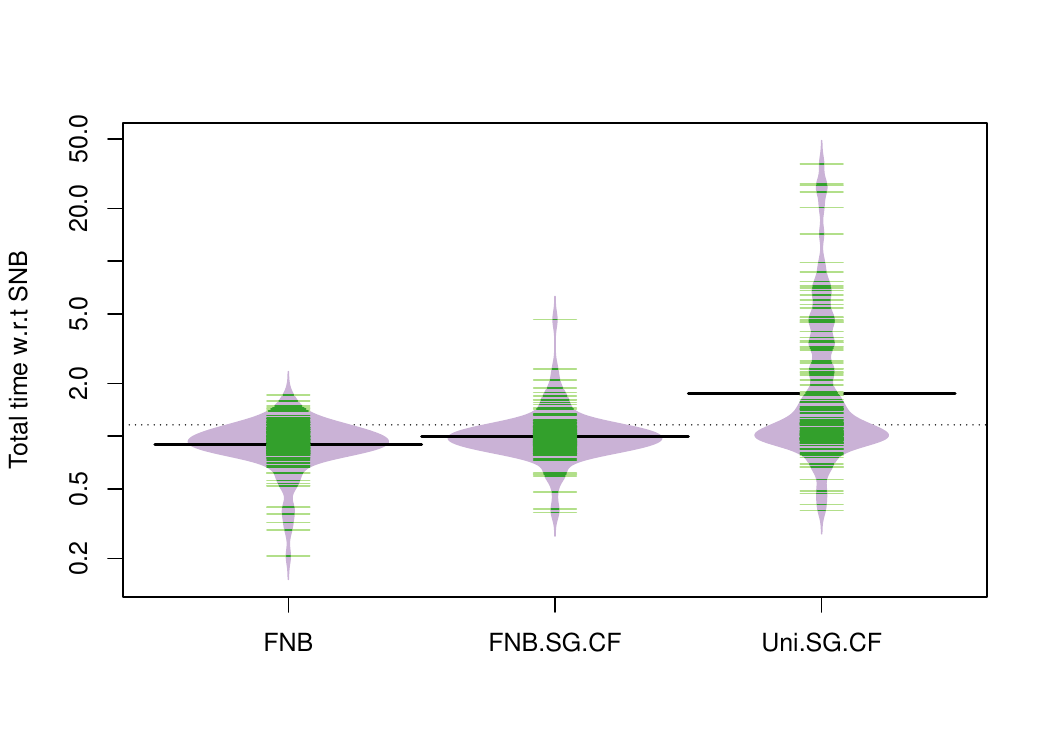}
\end{center}
\caption{Final comparison of the optimization performance w.r.t. the SNB method}
\label{FNBBestWrtSNB}
\end{figure}

\begin{figure}[!tbp]
\begin{center}
\includegraphics[width=0.65\textwidth, keepaspectratio]{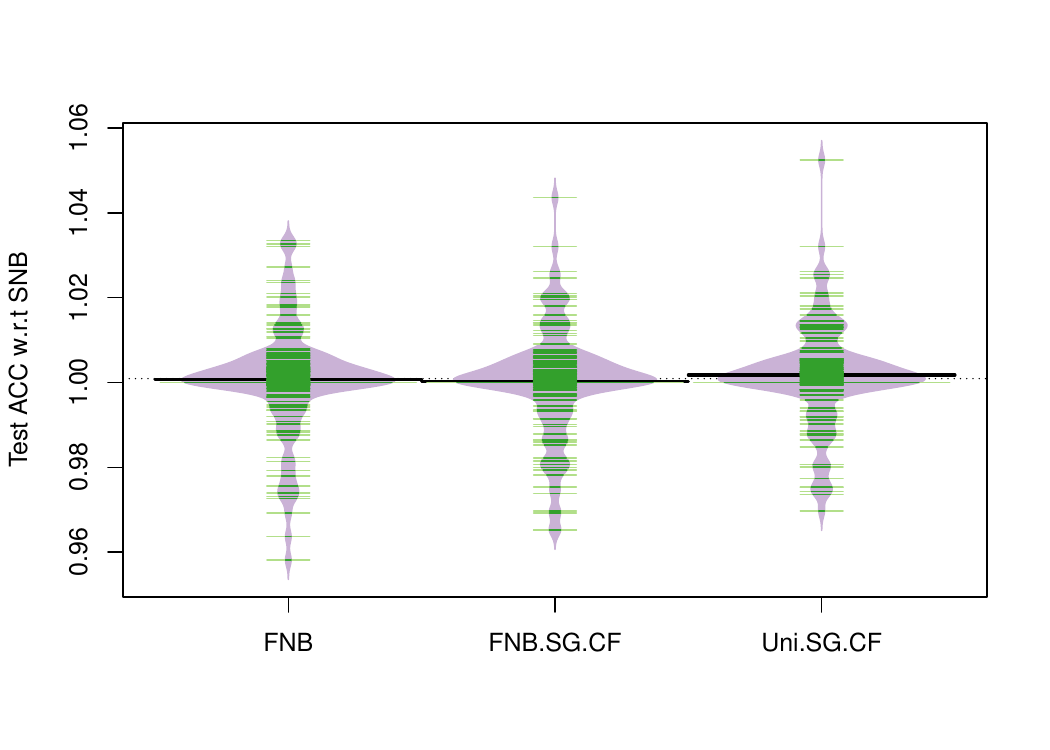}
\includegraphics[width=0.65\textwidth, keepaspectratio]{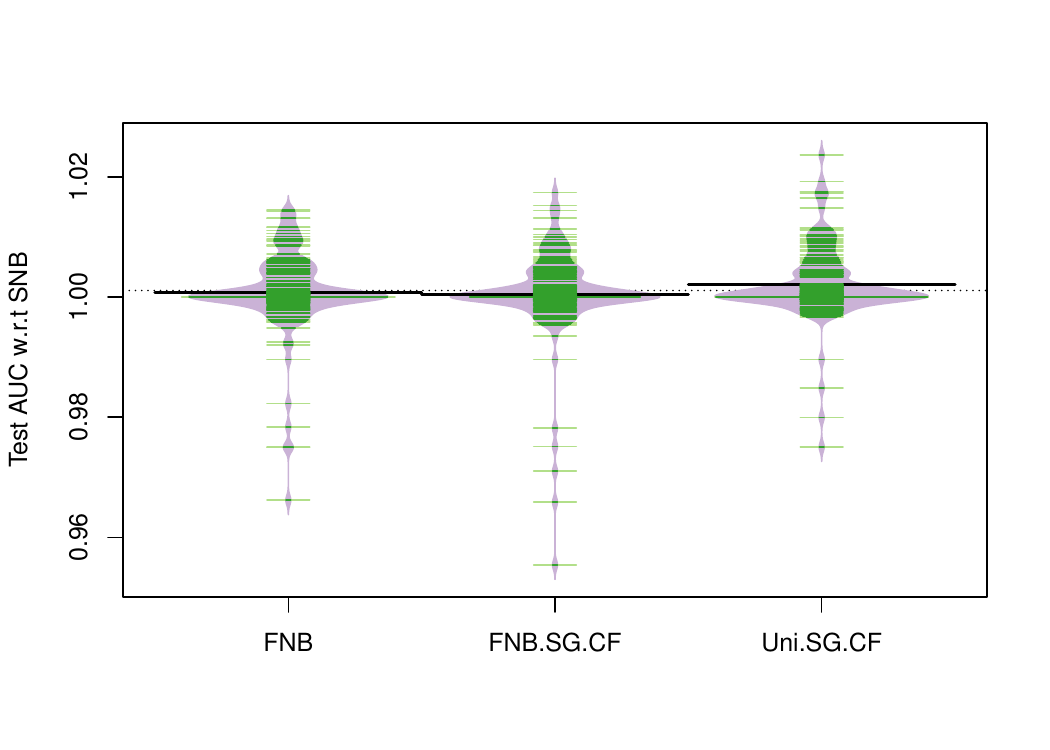}
\includegraphics[width=0.65\textwidth, keepaspectratio]{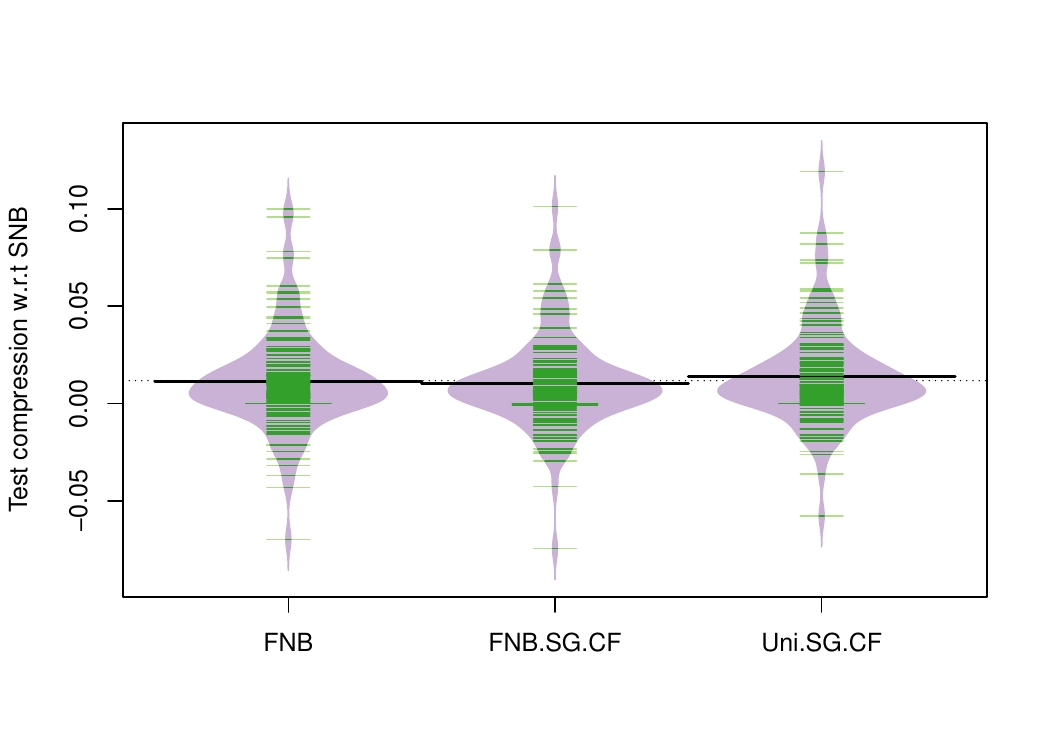}
\end{center}
\caption{Final comparison of the predictive performance w.r.t. the SNB method}
\label{FNBBestWrtSNB2}
\end{figure}

We finally study the performance of the FNB algorithm alone, not followed by SG.CF minimization. We compare it with the SG.CF initialized with the FNB and with the SG.CF with the uniform initialization. In Figure \ref{FNBBestWrtSNB} and \ref{FNBBestWrtSNB2} we present several comparison criteria of these three algorithm  with respect to SNB performance in order to analyze which of them could best replace the SNB predictor. Concerning the final train criterion on top of Figure \ref{FNBBestWrtSNB}, we can see that the three algorithms have the same quality of optimization. The ratios are computed by evaluating the value of the continuous criterion using $f_C$ regularization for all algorithms. The predictive performance plots on Figure \ref{FNBBestWrtSNB2} show a slight superiority of the SG.CF algorithm with uniform initialization but the predictive performance of the FNB alone are as good as those of the SNB. Concerning parsimony, the FNB algorithm alone is as parsimonious as the FNB algorithm followed by the SG.CF optimization and both are more parsimonious than the SG.CF optimization after uniform initialization. In any case, the number of variables is greatly reduced compared to the number of variables obtained with the SNB algorithm. Concerning optimization time, the FNB algorithm alone takes the same time as the SNB and less time than the SG.CF algorithms.  At last, in our software engineering context where the code maintainability is crucial, this fractional naïve Bayes alone is much easier to implement and maintain than the SG.CF algorithm. We then decide to replace in our software the SNB algorithm by the Fractional Naive Bayes algorithm.

\section{Conclusion}

\begin{figure}[!tbp]
\begin{center}
\includegraphics[width=0.49\textwidth, keepaspectratio]{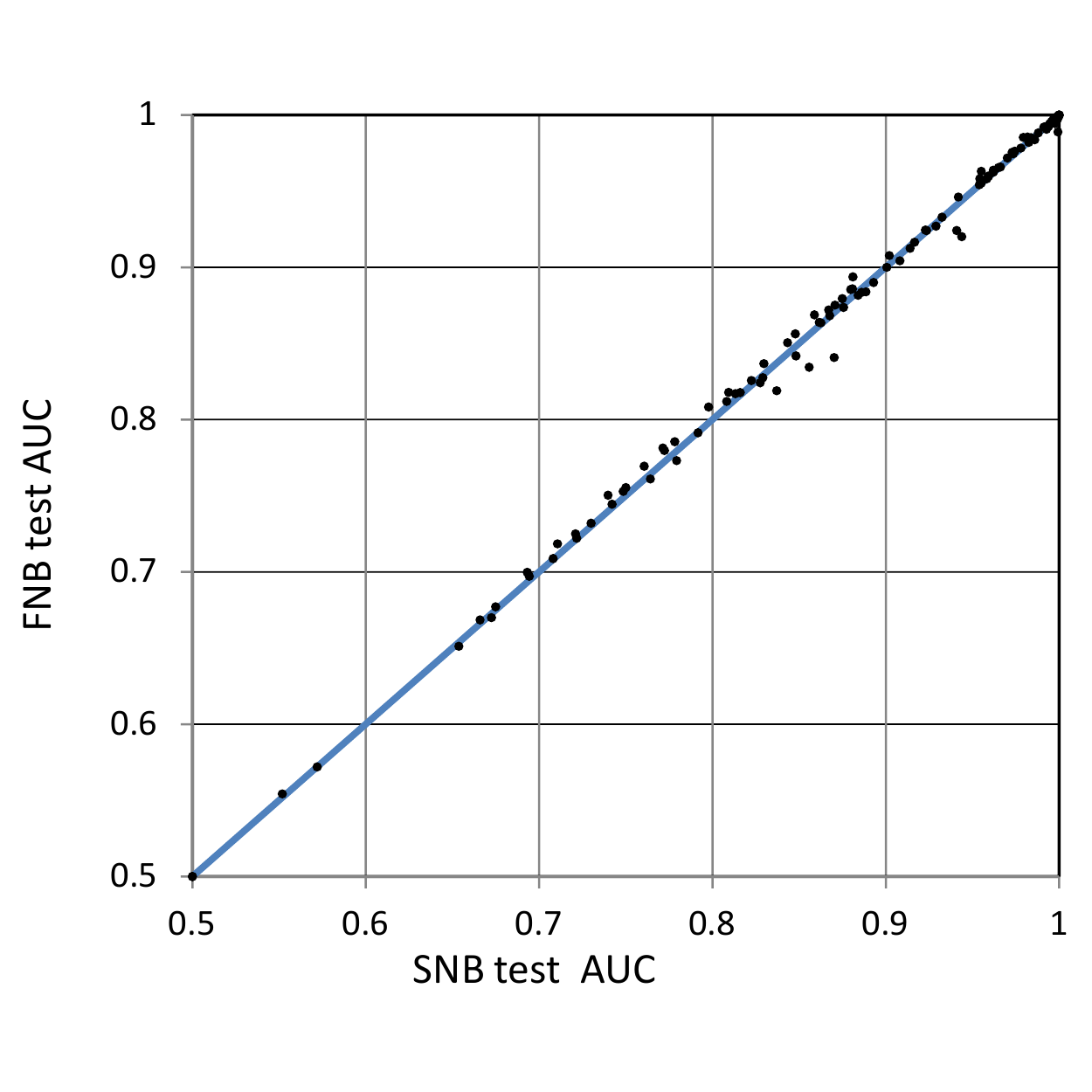}
\includegraphics[width=0.49\textwidth, keepaspectratio]{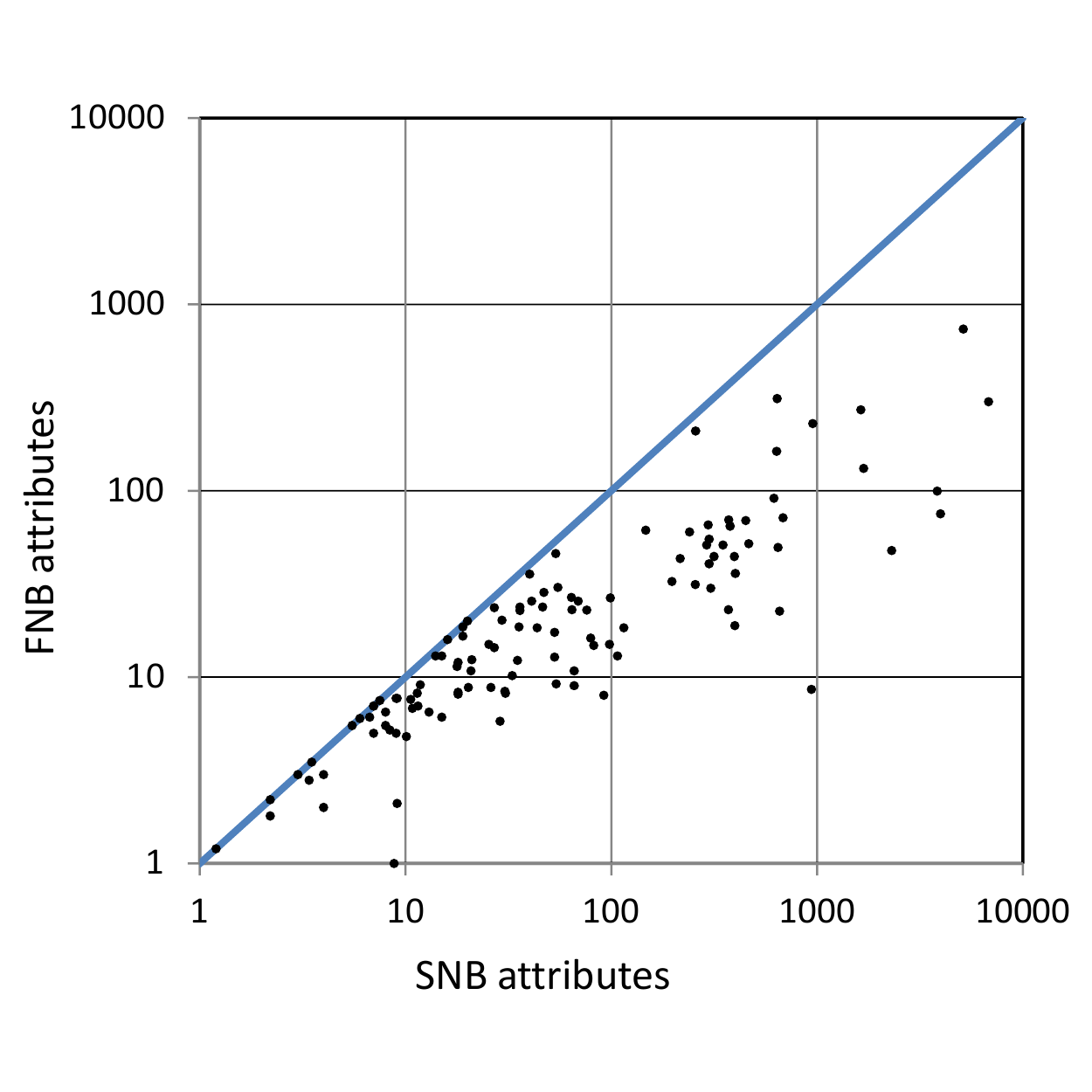}
\end{center}
\caption{FNB versus SNB: test AUC and number of attributes}
\label{FinalFNBvsSNB}
\end{figure}

We have studied an extension of the selective naive Bayes classifier based on a direct optimization of the variable weights. Our main expectation was to obtain parsimonious robust models with less variables and equivalent accuracy, compared to the SNB predictor used up until now, where the variable weights come from model averaging rather than from optimization.
We have proposed a sparse regularization of the model log-likelihood which takes into account prior regularization costs relative to each input variable.
The direct estimation of the variable weights comes down to a non-convex optimization problem for which we have proposed several algorithms. 
The experiments have shown that the best algorithm is a two-stage algorithm that exploits a simple gradient method during the first stage and a local optimization using composite function in the second stage.
Sensitivity analysis have been performed to parametrize the selected algorithm. To reduce the optimization time obtained with uniform initialization, two other initializations have been evaluated. The one obtained using a Fractional Naïve Bayes (FNB) results in the best trade-off between predictive performance, sparsity and optimization time.
In a software engineering context where code maintainability is crucial, this Fractional Naïve Bayes predictor alone provides an easy-to-maintain parsimonious model with good predictive performance and excellent sparsity. This is illustrated in the synthetic Figure~\ref{FinalFNBvsSNB}: we can see that the predictive performance evaluated using the test AUC is almost the same for both methods, while the sparsity of the solution is far better with the FNB method. Our comparison with the best algorithm issued from non convex optimization guarantees us a high level performance quality for this predictor.
It is implemented in the last version of the AutoML Khiops tools \cite{Khiops10}.

\begin{acknowledgements}
The authors would like to thank Yurii Nesterov who helped analyzing the penalized criterion and who proposed optimization algorithms adapted to the problem.
\end{acknowledgements}

\bibliographystyle{spmpsci}      
\bibliography{ArticleBilanDNB}

\end{document}